\documentclass[]{youtu} 

\usepackage{mathpazo}
\usepackage{graphicx}
\usepackage[numbers]{natbib}
\usepackage{threeparttable}
\usepackage{adjustbox}
\usepackage{amsmath}
\usepackage{subcaption}
\usepackage{tcolorbox}
\usepackage{colortbl}
\definecolor{forward}{RGB}{84, 130, 53}
\definecolor{inverse}{RGB}{47, 85, 151}
\definecolor{resist}{RGB}{128, 0, 128}
\definecolor{rebound}{RGB}{133, 19, 33}
\usepackage{xcolor}

\definecolor{c1}{RGB}{249,242,234}
\definecolor{c2}{RGB}{228,246,246}
\definecolor{c3}{RGB}{223,243,230}
\definecolor{c4}{RGB}{224,222,241}
\usepackage{multirow}
\usepackage{CJK}

\definecolor{blue(ryb)}{rgb}{0.01, 0.28, 1.0} 
\definecolor{azure(colorwheel)}{rgb}{0.0, 0.5, 1.0}
\definecolor{turquoise}{rgb}{0.19, 0.84, 0.78}

\definecolor{def}{RGB}{119, 228, 200}
\definecolor{thm}{RGB}{69, 53, 193}

\newtheorem{principle}{Principle}[section]

\newtcolorbox{thmbox}[1][]{colback=thm!5!white,colframe=thm!60!black,boxsep=-4pt,grow to left by=4pt,left=10pt,grow to right by=4pt,right=10pt,top=10pt,bottom=10pt,#1}
\newtcolorbox{defbox}[1][]{colback=def!5!white,colframe=def!60!black,boxsep=-4pt,grow to left by=4pt,left=10pt,grow to right by=4pt,right=10pt,top=10pt,bottom=10pt,#1}

\title{SmartSnap: Proactive Evidence Seeking for \\Self-Verifying Agents}
\author{Youtu-Agent Team$^*$}

\date{December 26, 2025}
\correspondence{caishaofei@stu.pku.edu.cn; yuleiqin@tencent.com}
\sourcecode{https://github.com/TencentYoutuResearch/SmartSnap}
\data{https://huggingface.co/collections/yolay/smartsnap}

\begin{document}

\abstract{
Agentic reinforcement learning (RL) holds great promise for the development of autonomous agents under complex GUI tasks,
but its scalability remains severely hampered by the verification of task completion.
Existing task verification is treated as a passive, post-hoc process: a verifier (i.e., rule-based scoring script, reward or critic model, and LLM-as-a-Judge) analyzes the agent's entire interaction trajectory to determine if the agent succeeds.
Such processing of verbose context that contains irrelevant, noisy history poses challenges to the verification protocols and therefore leads to prohibitive cost and low reliability.
To overcome this bottleneck, we propose \textbf{SmartSnap}, a paradigm shift from this passive, post-hoc verification to proactive, in-situ self-verification by the agent itself.
We introduce the \textbf{Self-Verifying Agent}, a new type of agent designed with dual missions: to not only complete a task but also to prove its accomplishment with curated snapshot evidences.
Guided by our proposed \textbf{3C Principles} (Completeness, Conciseness, and Creativity),
the agent leverages its accessibility to the online environment to perform self-verification on a minimal, decisive set of snapshots.
Such evidences are provided as the sole materials for a general LLM-as-a-Judge verifier to determine their validity and relevance.
Experiments on mobile tasks across model families and scales demonstrate that our SmartSnap paradigm allows training LLM-driven agents in a scalable manner,
bringing performance gains up to 26.08\% and 16.66\% respectively to 8B and 30B models.
The synergizing between solution finding and evidence seeking facilitates the cultivation of efficient, self-verifying agents with competitive performance against DeepSeek V3.1 and Qwen3-235B-A22B.
}
\maketitle

\renewcommand{\thefootnote}{*}
\footnotetext{Full author list in contributions.}
\renewcommand{\thefootnote}{\arabic{footnote}}


\section{Introduction}
\label{sec:intro}

\begin{figure}[htbp]
\begin{center}
\includegraphics[width=0.45\linewidth]{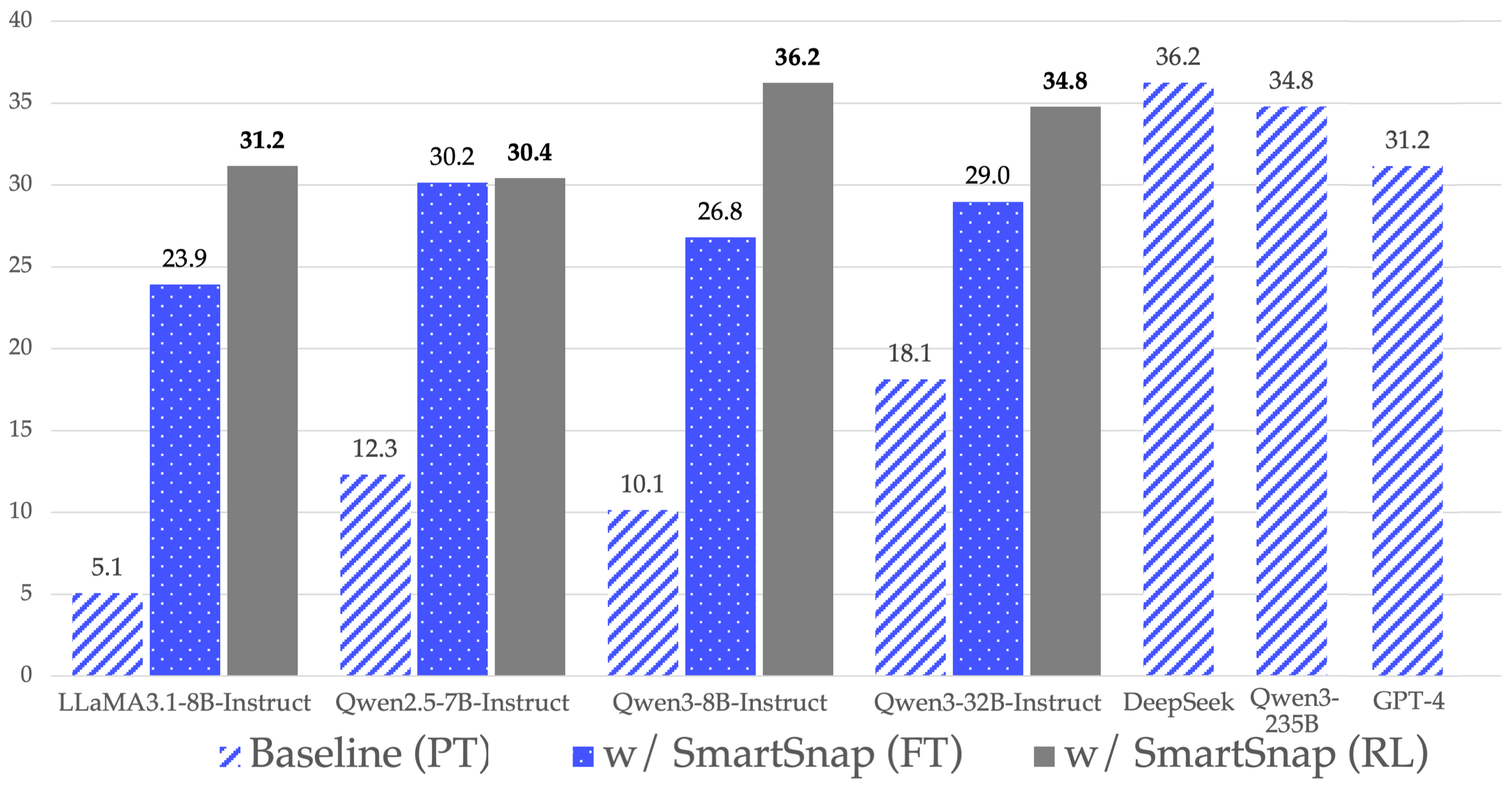}
\end{center}
\caption{Success rate on AndroidLab~\cite{xu2024androidlab} across model families and scales.
Compared with the vanilla prompting (PT),
our SmartSnap brings significant gains via fine-tuning (FT) and reinforcement learning (RL) without relying on sophisticated rule-based verifiers and task-specific reward models.
The developed self-verifying agents learn to complete tasks and curate snapshot evidences in a complementary manner,
achieving competitive performance with larger LLMs.} 
\label{fig:androidlabintro}
\end{figure}

\begin{figure}[t]
\begin{center}
\includegraphics[width=0.9\linewidth]{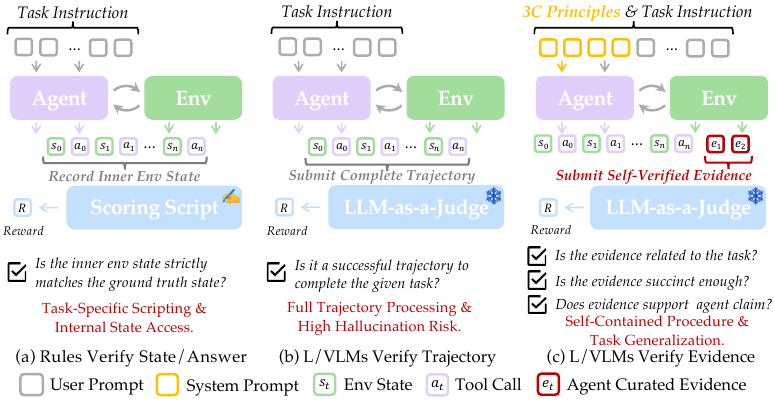}
\end{center}
\caption{Three strategies for agent verification distinguished by their inputs to the verifier: 
(a) the task-specific script accessing the \textit{ground-truth state}; 
(b) the \textit{full trajectory with noisy context};
and (c) the \textit{agent-curated evidence set}.
}
\label{fig:comparison}
\end{figure}

The recent rapid advancement of Large Language Models (LLMs)~\citep{gpt3, guo2025deepseek, qwen, gemini} has catalyzed the emergence of general-purpose agents capable of understanding human intent and operating complex digital interfaces.
The vision is to automate daily tasks on devices like computers~\citep{ufo2, qin2025ui, xie2024osworld} and smartphones~\citep{ye2025mobile,xu2024androidlab,androidworld,xu2025mobilerlonlineagenticreinforcement}, promising a significant boost in human productivity. The challenge of scaling the training of such agents, a pursuit now often termed Agentic Reinforcement Learning (Agentic RL)~\citep{agenticRL,zhang2025agentrlscalingagenticreinforcement}, critically hinges on a scalable and automated verification mechanism~\cite{shao2025deepseekmathv2selfverifiablemathematicalreasoning}.
In traditional, constrained environments such as video games~\citep{atari, moba}, verification is unambiguous; success is determined either by explicit reward signals like a game score, or through rule-based scripts that directly access the environment's internal state for validation (as depicted in Figure \ref{fig:comparison}a). Conversely, for agents in open-ended digital worlds, the spectrum of tasks is vast, and their success criteria are not amenable to simple manual specification.
Although early studies on open-ended tasks attempted to use manually designed, rule-based scripts to verify task completion~\citep{uir1, liu2025infigui},
these approaches remain labor-intensive and unscalable towards diverse real-world tasks.

To overcome this problem, recent research has shifted towards employing Vision-Language Models (VLMs)~\citep{qwen25vl} as automated verifiers~\citep{yang2025zerogui, opencua, qi2024webrl, autoglm} (Figure \ref{fig:comparison}b).
This paradigm offers immense scalability and generalization, as a single VLM can, in principle, verify a wide range of tasks without per-task engineering.
However, this shift exposes a deeper, architectural flaw in the prevailing agent paradigm: \textit{the agent's execution is entirely decoupled from the subsequent verification process, reducing verification to a difficult, passive, and post-hoc analysis of potentially ambiguous behaviors}.
In this paradigm, the agent itself is \textbf{\textit{verification-agnostic}}; it performs actions without any consideration of proving its success beforehand.
Consequently, the full cognitive burden of verification falls upon the VLM verifier, which is forced to review the entire, often noisy and ambiguous, interaction trajectory in a brute-force manner.
Such passive verification design leads directly to two critical drawbacks:
(1) it incurs prohibitively high verification costs (in both API fees and latency),
and (2) it places a heavy load on the VLMs, increasing the risk of hallucinations and false positive judgments~\citep{yang2025zerogui}.

We propose SmartSnap, a paradigm shift from the flawed paradigm of \textit{passive external verification} to \textit{proactive self-verification} for building GUI agents.
Specifically, we redesign the core responsibilities of the agent itself and introduce the concept of a \textit{\textbf{Self-Verifying Agent}}. 
It fulfills two-fold missions: to \textit{Execute} and to \textit{Verify}.
Upon completing a task, instead of passively awaiting judgment, the agent leverages its privileged position within the live environment sandbox (that is often destroyed or unavailable by the time of post-hoc verification).
It actively reviews its interaction history and, crucially, can creatively execute additional, evidence-oriented actions to generate more decisive proof.
This entire curation process, guided by what we term the \textit{\textbf{3C Principles of Evidence Curation}} (Completeness, Conciseness, and Creativity), yields a concise yet decisive set of evidence.
We posit that this dual mission creates a \textit{\textbf{synergistic learning loop}}: the requirement to find and present evidence compels the agent to develop a deeper, more holistic understanding of the task, as the core knowledge and capabilities required for successful execution and effective proof are deeply intertwined.
Through this proactive self-verification,
the inputs to the VLM verifier are reduced from the full trajectory to a minimal, curated evidence set, fundamentally resolving the trade-off between verification cost and reliability.
Beyond delivering a final verdict, we equip the verifier to provide fine-grained feedback on evidence quality, facilitating a reward-shaping process that accelerates the agent's acquisition of evidence curation skills. 
As shown in Figure~\ref{fig:androidlabintro},
both fine-tuning (FT) and RL under our SmartSnap paradigm lead to significant performance gains on various LLM families and scales.

Our main contributions are threefold:
\begin{itemize}
    \item We propose the \textit{\textbf{SmartSnap}}, a new paradigm that shifts the burden of verification from the external handcrafted verifier to the \textit{\textbf{Self-Verifying Agent}}. This fundamentally enhances the efficiency, scalability, and robustness of the verification process by reducing the cognitive load of verifier under GUI tasks. 
    
    \item We formalize the \textit{\textbf{3C Principles}} of evidence curation (\textit{Completeness, Conciseness, and Creativity}), providing a theoretical foundation for training self-verifying agents. 
    
    \item We design an efficient learning framework where an LLM/VLM verifier provides structured, dense feedback. This is achieved through intrinsic \textit{\textbf{Reward Shaping}}, not only to judge task completion but also to explicitly guide the agent in improving the quality of its evidence curation.
\end{itemize}

\section{Related Works and Preliminaries}
\label{sec:preliminaries}

\subsection{Agents based on Large Language Models}
Agents based on LLMs~\citep{guo2025deepseek, gemini, qwen} are entities that leverage the world knowledge acquired from web-scale pre-training and RL post-training to perform instruction understanding, perception, task planning, and decision-making. A prominent line of research focuses on enabling these agents to operate general-purpose electronic devices, such as computers~\citep{qin2025ui, xie2024osworld} and smartphones~\citep{ye2025mobile, xu2024androidlab, androidworld}, to fulfill users' daily instructions or to play games~\citep{wang2023describe, cai2024groot, cai2025rocket, voyager}. Existing approaches can be broadly categorized into two paradigms based on their training methodology: \textbf{Supervised Fine-Tuning (SFT)} and \textbf{Reinforcement Learning (RL)}. 
SFT-based methods~\citep{lu2024gui, chen2024guicourse, qin2025ui} typically rely on expert demonstration trajectories recorded by human annotators for learning. 
Distillation~\cite{cui2024sinkhorn,cui2025multi,cui2025optical} from stronger LLMs as teacher also provides effective supervision.
Nevertheless, a primary limitation of this approach is its low data collection efficiency and high cost. To improve data utility, UI-TARS-2~\citep{wang2025ui} employs a rejection sampling strategy, where successful trajectories from online interaction with the environment are automatically filtered and used for iterative fine-tuning. Nevertheless, these methods still struggle to eliminate the dependency on a large corpus of high-quality, successful trajectories. 
To further reduce human intervention, researchers~\citep{ye2025mobile, liu2025infigui,qi2024webrl,autoglm,qin2025learnropestrustwins} have turned to the RL paradigm, which enables learning from reward signals. However, the core challenge for RL-based methods lies in designing the reward function. Early works, such as UI-R1~\citep{uir1} and InfiGUI-R1~\citep{liu2025infigui}, relied on handcrafted, rule-based reward functions; however, this approach proved to be labor-intensive, unscalable, and prone to overfitting, thereby lacking generalization. To overcome these limitations, subsequent works like WebRL~\citep{qi2024webrl} and AutoGLM~\citep{autoglm} proposed training an Outcome Reward Model (ORM) to automatically judge if the final environmental state corresponds to task completion, thereby avoiding per-task rule design. More recently, approaches such as Zero-GUI~\citep{yang2025zerogui} have leveraged the power of VLMs to review all screenshots captured during a task for a comprehensive judgment. While this method cleverly mitigates agent hallucination and reduces the false positive rate through mechanisms like voting, the prohibitively high cost of full-trajectory review and the extreme demands on the VLM's long-range reasoning capabilities remain significant drawbacks.



\subsection{Group Relative Policy Optimization}
Inspired by recent works \citep{shao2024deepseekmath, guo2025deepseek}, we adopt \textit{Group Relative Policy Optimization (GRPO)} as our learning algorithm. GRPO is a policy gradient method that innovatively eliminates the need for a separate critic network. Its key mechanism is to compute the advantage function by normalizing the reward of each trajectory against the statistics (mean and standard deviation) of a group of trajectories sampled from the same policy. This approach significantly reduces the training cost and memory overhead associated with traditional actor-critic methods. For each task instruction $q$, the training process begins by sampling a group of $G$ trajectories, $\{\tau_1, \dots, \tau_G\}$, with rewards, $\{R_1, \dots, R_G\}$, from the current policy $\pi_{\theta_\text{old}}$. The policy parameters $\theta$ are then updated by maximizing the following objective function: 
\begin{equation}
    \begin{split}
    \mathcal{J} = \frac{1}{G}\sum\limits_{i=1}^{G} \min\left(r_i(\theta)A_i,\text{clip}\left( r_i(\theta), 1-\epsilon, 1+\epsilon \right)A_i \right), \\
    r_i(\theta) = \frac{\pi_\theta(\tau_i|q)}{\pi_{\theta_\text{old}}(\tau_i|q)},\ \ A_i = \frac{R_i - \text{mean}(\{R_1, \cdots, R_G\})}{\text{std}(\{R_1, \cdots, R_G\})}. 
    \end{split}
\end{equation}

\section{Method}
\label{sec:method}

\subsection{Overall Pipeline and Problem Formulation} 
We formalize the process of a self-verifying agent's task execution and evidence curation within an interactive environment as an augmented Markov Decision Process (MDP). 
We model the environment as a tuple $\mathcal{M} = (\mathcal{S}, \mathcal{A}', P)$. 
$\mathcal{A}'$ is an \textbf{augmented action space}, defined as the union of two distinct sets of actions: $\mathcal{A}' = \mathcal{A}_{\text{exec}} \cup \mathcal{A}_{\text{curate}}$, where 
$\mathcal{A}_{\text{exec}}$ denotes the actions for direct environment interaction (e.g., \texttt{click(x, y)}, \texttt{type('text')}, \texttt{screenshot()}). 
A state $s \in \mathcal{S}$ can represent a snapshot of the environment or a tool response. 
 $P(s_{t+1} | s_t, a_t)$ is the state transition function, where $a_t \in \mathcal{A}_{\text{exec}}$. 
$\mathcal{A}_{\text{curate}}$ is the set of actions used for submitting proof of task completion. 
A task is defined by a language sentence $I$. The agent's behavior is governed by a unified policy $\pi_\theta(a_t | s_{\le t}, a_{< t}, I)$, which maps the history observations and the instruction $I$ to a probability distribution over the augmented action space $\mathcal{A}'$. 
By interacting with the environment, the policy $\pi_\theta$ generates a trajectory $\tau = (s_0, a_0, s_1, \dots, s_N, a_N)$. The trajectory terminates when the agent selects a special terminal action $a_N \in \mathcal{A}_{\text{curate}}$, such as \texttt{submit('text',E)}.
The \texttt{'text'} denotes the final message $T$ and $E$ refers to the curated evidence set.
We can use any state-of-the-art LLMs as the verifier $\psi$. It acts as a scoring function which takes the instruction $I$, text message $T$, the evidence $E$, and evaluation guidance $G$ as input, returning a reward $R = \psi(G, I, T, E)$.

\subsection{Grounding Evidence in Tool Interactions}



A central challenge in our SmartSnap framework is defining trustworthy evidence.
A naive approach is relying on the agent's self-declaration, where the agent generates a textual summary like, \textit{``Based on the instruction, I performed actions X and Y, observed Z, and thus the task is complete.''}
This is inherently untrustworthy, as the agent is incentivized to omit failures to \textbf{hack} the verifier.
A slightly more robust method is using the final state screenshot.
However, this approach is also inherently brittle for two key reasons.
First, the final state is not always informative; for instance, an agent might successfully fill a table but then navigate back to a generic home page before concluding, leaving the final screenshot useless.
Second, verifying certain tasks requires capturing the difference before and after a critical action (e.g., filling the correct arguments into the app as required), where a single, static screenshot is insufficient to prove. 
To overcome these limitations, we formally define a single piece of evidence, an exhibit, as an atomic tuple representing a direct cause-and-effect event: $(a_t, s_{t+1})$
where $a_t$ is the action (i.e., tool call) executed by the agent, and $s_{t+1}$ is the direct observation (i.e., tool response). This interaction pair is an objective, unalterable fact rather than the agent's subjective summary.
This atomic definition elegantly simplifies the agent's task: \textit{instead of generating descriptive text, the agent learns to output a set of integer indices $E$.}
These indices pinpoint the most critical interactions within its full history, and this curated evidence is then programmatically formatted using the standard OpenAI chat template:
\begin{equation}
\text{StringJoin} \left(\{ \mathcal{F}(a_e, s_{e+1}) | e \in E\}\right),
\end{equation}
where $\text{StringJoin}(\cdot)$ denotes the pythonic concatenation operation of string and $\mathcal{F}$ represents the OpenAI chat template formatting operation. This interaction tuple links a specific action to its immediate, multimodal outcome, and therefore, the agent's core learning objective becomes the selection of a minimal yet decisive evidence set $E$ from its entire interaction history.

\subsection{Foundational Principles of Evidence Curation}
To guide the agent's evidence curation behavior, we establish the \textbf{3C Principles} that govern it: \textit{Completeness}, \textit{Conciseness}, and \textit{Creativity}. These principles are formulated as meta-instructions and injected into the agent's system prompt to constrain its decision-making process. 
\begin{defbox}
\begin{principle}[Completeness and Conciseness]
The ideal evidence set should maximize the signal of proof while minimizing the noise of redundancy. 
\end{principle}
\end{defbox}
Operationally, the system prompt drives the agent to review its entire history of tool interactions after task execution.
The \textit{Completeness} principle requires the agent to ensure that all pivotal tool interactions are included within the final evidence set $E$. From a statistical perspective, the goal is to \textbf{maximize the True Positive (TP) rate} of the verifier. This prevents a correctly executed trajectory from being wrongly penalized due to incomplete evidence. 
Conversely, the \textit{Conciseness} principle is crucial for \textbf{minimizing the False Positive (FP) rate}. 
This pursuit stems from a dual consideration of efficiency and robustness. For an LLM/VLM verifier, a shorter, more information-dense submission reduces its cognitive load and the likelihood of hallucination induced by irrelevant details~\citep{yang2025zerogui}. 
From another perspective, avoiding excessively long inputs for the verifier mitigates the risk of \textbf{context rot}~\citep{context-rot} (i.e., the degradation of the LLM verifier's performance as the context length increases). 
\begin{figure}[t]
\begin{center}
\includegraphics[width=0.6\linewidth]{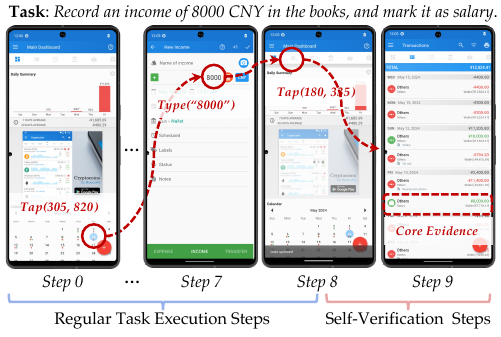}
\end{center}
\caption{
An example of self-verification for evidence curation.
The agent decomposes the task into an actionable checklist where the date, the amount, and the category tag are to be confirmed during stepwise task completion.
The proactive step of taking snapshots that list the target transaction provides a definitive evidence for task completion.
}
\label{fig:infer_pipe}
\end{figure}

\begin{defbox}
\begin{principle}[Creativity]
If the necessary evidence does not exist in the interaction history, the agent is encouraged to create it by taking additional actions. 
\end{principle}
\end{defbox}
This principle elevates the agent from a passive historian of its actions to a proactive investigator. It establishes a crucial distinction between two types of agent behavior: task-oriented actions, which are taken to achieve the task's goal, and evidence-oriented actions, which may be taken after the goal is met, with the sole intent of generating a better proof. 
For instance, consider the task of \texttt{installing a browser extension}. The final task-oriented action is clicking the ``Add extension'' button in a confirmation dialog. The immediate feedback for this action---the dialog disappearing---does not actually confirm that the extension has been successfully loaded into the browser, thus constituting weak evidence. A creative agent, after performing this click, will shift its attention from the web content to the browser's own user interface, proactively inspecting the ``toolbar area'' for the new extension's icon. This inspection forms a perfect evidence-oriented action. 
Such capability to create evidence is the key to achieving true robustness in agentic RL:
1) It allows the agent to understand the \textbf{indirect consequences} of its actions and to search for an optimal proof within a broader state space that includes the entire sandbox (e.g., a complete application UI), thereby satisfying both the completeness and conciseness criteria.
2) It encourages the agent to focus on the state transition and get acquainted with the environment feedback.
3) It implicitly incentivizes reasoning and reflection of agent for active collection of evidence according to the environment feedback.

\subsection{Evidence Verification and Reward Shaping}
To provide a dense and informative reward signal that guides the agent in mastering the complex evidence curation process, we designed a principled verification framework. This process is executed by an LLM/VLM Verifier following detailed instructions, and its structured feedback is used to construct a multi-component reward function. 

\noindent \textbf{Evidence Validity Check.}
As the first stage of verification, the verifier determines if the agent's submitted evidence is relevant to the task instruction. Critically, evidence that unambiguously proves \textit{failure} (e.g., a screenshot showing Wi-Fi is still on for a ``turn off Wi-Fi'' task) is also considered valid. Evidence is only deemed invalid if it is entirely irrelevant. We provide a small positive reward $R_{\text{validity}}$ (e.g., $+0.5$) for valid evidence and a penalty otherwise. This auxiliary signal encourages the agent to navigate towards relevant application screens, reinforcing its ability to identify key steps even if the task ultimately fails. 

\noindent \textbf{Strict Evidence-to-Task Grounding.} 
Only when the agent claims success does the verifier proceed to this core stage of scrutiny. This stage adheres to a ``success only upon unequivocal proof'' rule, conducting a rigorous, multi-faceted examination of the evidence. This includes checks for (1) Zero Assumptions: The Verifier must not infer or ``fill in the gaps'' for any states or actions that are not explicitly and unambiguously present in the evidence. What is not shown in the evidence is assumed not to have happened. 
(2) Traceable Reasoning: Every factual claim made by the Verifier in its analysis must be directly traceable to a specific piece of evidence (e.g., ``In the screenshot from Round 3...''). This mechanism promotes literal description over potentially hallucinatory summarization. This stage results in a outcome reward $R_{\text{complete}} \in \{1, 0\}$.

\noindent \textbf{Behavioral and Formatting Regularization.}
To further refine the agent's curation behavior, we introduce two additional auxiliary reward components:
(1) Format Reward ($R_{\text{format}}$). This signal constrains the agent's output to adhere to the format specified in its system prompt strictly. Any formatting error incurs a fixed negative penalty, which is crucial for stable interaction with downstream modules.  
(2) Conciseness Penalty ($R_{\text{concise}}$). To enforce the Principle of Conciseness, we include a penalty term proportional to the size of the evidence, such as $R_{\text{concise}} = -\lambda \cdot \text{size}(E)$, where $\lambda$ is a small coefficient. This incentivizes the agent to find the most succinct proof, given that completeness is met.  

Ultimately, these four components form a composite reward function, $R_{\text{total}}$, which provides rich and dense feedback for the agent's end-to-end learning. The final reward is a sum of these components:
\begin{equation}
R = R_{\text{format}} + R_{\text{validity}} + R_{\text{complete}} + R_{\text{concise}}
\end{equation}
We optimize the policy using GRPO~\citep{shao2024deepseekmath} under the VeRL~\citep{verl} implementation.


\section{Experiment}
\label{sec:experiment}

\subsection{AndroidLab} 

\paragraph{Environments.}
We perform experiments on the AndroidLab~\citep{xu2024androidlab} benchmark, a reproducible evaluation platform featuring 138 tasks across nine apps, which run on predefined Android Virtual Devices (AVDs). For our work, we utilize only the defined tasks and their corresponding initial environments. We do not use AndroidLab's built-in, rule-based evaluation system, which determines task completion by decomposing tasks into sub-goals and matching specific UI tree elements. Avoiding this rule-based reward function is a key advantage of our Self-Verifying paradigm, as we believe designing such functions incurs a prohibitively high manual cost. 
Furthermore, our observation utilizes the compressed XML tree representation provided by Androidlab, rather than high-resolution screenshots. This choice allows us to focus on the agent's planning and decision-making capabilities, rather than being limited by insufficient perceptual abilities. Our action space adopts the native action space provided by Androidlab, which includes \texttt{Tap}, \texttt{Swipe}, \texttt{Type}, \texttt{Long Press}, \texttt{Home}, and \texttt{Back}. We additionally add a \texttt{submit} tool to allow the agent to present the evidence snapshot.

\begin{table}[htbp]
\centering
\caption{Statistics on the distribution of tasks from AndroidLab~\cite{xu2024androidlab}.}
\label{tab:androidlabdistribution}
\resizebox{\linewidth}{!}{
\setlength{\tabcolsep}{1mm}{
\fontsize{9pt}{10pt}\selectfont{
\begin{tabular}{lllllllllll}
\toprule
\multirow{2}{*}{Split} & \multicolumn{10}{c}{Number of Tasks (Percentage) per App Category} \\
 & Calendar & Zoom & Bluecoins & PiMusic & Maps.me & Contacts & Cantook & Clock & Setting  & Total \\
\midrule
Training & 92 (12.67\%) & 0 (0.00\%) & 81 (11.15\%) & 58 (7.98\%) & 86 (11.84\%) & 55 (7.57\%) & 84 (11.57\%) & 43 (5.92\%) & 227 (31.26\%) &  726 (100\%) \\
Validation  & 14 (10.14\%) & 5 (3.62\%) & 15 (10.86\%) & 12 (8.69\%) & 15 (10.86\%) & 15 (10.86\%) & 12 (8.69\%) & 27 (19.56\%) & 23 (16.66\%) &  138 (100\%) \\
\bottomrule
\end{tabular}
}}}
\end{table}

\paragraph{Datasets.}
To ensure comparability,
we prepare the same 726 training tasks as those released in Android Instruct dataset~\cite{xu2024androidlab}.
We use the same tasks for both cold-start SFT and RL.
Specifically for the SFT dataset,
it consists of two components:
1) We follow the VeRL agent loop~\cite{verl} to build a simple agent framework that allows the agent LLM to interact with the Android environment.
All the 726 training tasks are fed into the framework for trajectory generation where both task completion and evidence submission are performed.
To ensure diversity of the collected trajectories,
we adopt both the DeepSeek V3.1~\cite{liu2024deepseek} and the Qwen3-235B-A22B~\cite{yang2025qwen3} as agents for rollout.
2) We update the original response data of the Android Instruct dataset with those from DeepSeek V3.1 and Qwen3-235B-A22B.
In total,
we obtain around 550K trainable QA pairs from 30K trajectories.
We randomly sample 100K QA pairs as the cold-start dataset.
Table~\ref{tab:androidlabdistribution} provides the detailed statistics on the distribution of tasks in AndroidLab~\cite{xu2024androidlab}.

\subsection{Implementation Details}

\paragraph{Agent.}
In the present study,
we employ various baselines of different architectures, size, and edition as the backbone for Self-Verifying Agent:
LLaMA3.1-8B-Instruct~\citep{grattafiori2024llama},
Qwen2.5-7B-Instruct~\citep{qwen2025qwen25technicalreport},
Qwen3-8B-Instruct~\citep{yang2025qwen3},
and Qwen3-32B-Instruct~\citep{yang2025qwen3}.
It is noted that due to the limit of context length, we use the instruct mode for Qwen3 series rather than the verbose reasoning mode.
The selection of backbone models (e.g., size) is based on the requirement that the model must possess instruction following capability for comprehension of the \textbf{3C principles} defined in the system prompt.
Furthermore, unlike the majority of computer use agents~\citep{ye2025mobile, qin2025ui, yang2025zerogui} that compress history into a fixed-length context,
we do not bring any agent scaffold into training and use the entire trajectory for end-to-end optimization.
Such design enjoys two benefits:
1) the Self-Verifying agent can review the complete context for task-oriented reflection and evidence preparation;
2) the independence of complicated memory management in existing scaffold frameworks enables broader compatibility and easy accessibility. 
We design a system prompt of approximately 4k tokens, which specifies the agent's required output format, as illustrated in Figure~\ref{fig:comparison}.

\paragraph{Verifier.} For the verifier, we utilize the DeepSeek-R1 model~\citep{guo2025deepseek}. Since the agent's evidence is also in the form of compressed XML, DeepSeek-R1, which lacks multimodal understanding but possesses strong reasoning capabilities, is highly suitable. The system prompt designed for the verifier is shown in Figure B. We implement a structured reward system: if the agent provides evidence relevant to task verification, a reward of $0.2$ is given. If, based on this, the agent successfully completes the task, an additional $0.8$ reward is granted. To mitigate verifier hallucinations, we use DeepSeek-R1 to evaluate the same evidence three times; the task is only considered complete if at least two evaluations pass. Additionally, if the agent's output does not adhere to the format prescribed by the system prompt, a fixed penalty of $-1.0$ is applied.

\paragraph{Training.} We use the VeRL~\cite{verl} for both cold-start SFT and the RL with SmartSnap.
For SFT experiments,
we set the training batch size as $32$, number of epochs as $5$, learning rate default as $1e-5$ with \texttt{cosine} decay, and maximum context length of $32K$.
For RL,
we set the training batch size as $8$,
ppo mini batch size as $8$,
group size as $8$,
KL coefficient $\beta=0$,
constant learning rate default as $1e-6$,
maximum number of interaction turns as $30$,
the total number of training steps as 180 (2 epochs),
and the maximum context length of 32K.
All experiments are conducted on 64 GPUs.

\paragraph{Evaluation}
For the validation of task completion under AndroidLab environment,
we follow~\cite{xu2024androidlab} to use the LLM-as-a-Judge (GLM4)~\cite{glm2024chatglm} for fair comparison.
It analyzes all the interaction trajectories (e.g., traces and \texttt{xml} snapshots) for final scoring.

\subsection{Main Results}


\begin{table}[htbp]
\centering
\caption{Results on AndroidLab with XML mode.
PT, FT, and RL stand for prompting, fine-tuning, and reinforcement learning, respectively.
SR, Sub-SR, RRR, and ROR stand for Success Rate, Sub-Goal Success Rate, Reversed Redundancy Ratio, and Reasonable Operation Ratio, respectively.
For all these metrics, a higher value means better performance.
The best and second-best results are marked in \textbf{bold} and \underline{underlined}, respectively.
We do not report RRR score if SR < 5.
}
\label{tab:androidlab}
\resizebox{0.6\linewidth}{!}{
\setlength{\tabcolsep}{1mm}{
\fontsize{9pt}{10pt}\selectfont{
\begin{threeparttable}
\begin{tabular}{llllll}
\toprule
Type & Model & SR & Sub-SR & RRR & ROR \\
\midrule
PT & GPT-4o & 25.36  & 30.56  & \textbf{107.45} & 86.56 \\
PT & GPT-4-1106-Preview & 31.16 & 38.21 & 66.34 & 86.24 \\
PT & Gemini-1.5-Pro & 18.84 &  22.40 & 57.72 &  83.99 \\
PT & Gemini-1.00 &  8.70 & 10.75 & 51.80 & 71.08  \\
PT & GLM4-Plus &  27.54 & 32.08 & 92.35 &  83.41  \\
PT & DeepSeek-V3.1 & \textbf{36.23}  &  \underline{40.95}  & 81.01  & 94.63  \\
PT & Qwen3-235B-A22B & \underline{34.78} & 38.76 & 83.35 & 89.48 \\
\midrule
\multicolumn{6}{c}{\texttt{Act}-only$^{*}$} \\
PT & LLaMA3.1-8B-Instruct$^{\ddagger}$ & 2.17 &  3.62 & -- &  52.77 \\
FT$^{\dagger}$~\cite{xu2024androidlab} & LLaMA3.1-8B-Instruct$^{\ddagger}$ & 23.91{\tiny \textbf{\textcolor{teal}{(+21.74\%)}}}  &  30.31 & 75.58 & 92.46 \\
\rowcolor[gray]{.95}
PT &  LLaMA3.1-8B-Instruct   & 5.07  &  6.28  &  52.77  & 51.82   \\
FT$^{\dagger}$~\cite{xu2024androidlab} & LLaMA3.1-8B-Instruct & 20.28{\tiny \textbf{\textcolor{teal}{(+15.21\%)}}}  & 26.13  & 69.44 & 90.43 \\
FT (ours) & LLaMA3.1-8B-Instruct & 23.91{\tiny \textbf{\textcolor{teal}{(+18.84\%)}}} & 30.36 & 37.96  & 83.23 \\
\rowcolor{c2}
RL (ours) & LLaMA3.1-8B-Instruct & 31.15{\tiny \textbf{\textcolor{teal}{(+26.08\%)}}} & 38.03 & 81.28 & \underline{95.80} \\
\multicolumn{6}{c}{\texttt{ReAct}} \\
\rowcolor[gray]{.95}
PT & Qwen2.5-7B-Instruct & 12.32 & 14.98 & 67.56 & 78.52 \\
FT$^{\dagger}$~\cite{xu2024androidlab} & Qwen2.5-7B-Instruct & 20.28{\tiny \textbf{\textcolor{teal}{(+7.96\%)}}} & 27.05 & 35.52 & 62.46 \\
FT (ours) & Qwen2.5-7B-Instruct & 30.15{\tiny \textbf{\textcolor{teal}{(+17.83\%)}}}  & 36.59  & 49.19  & 73.28 \\  
\rowcolor{c2}
RL (ours) & Qwen2.5-7B-Instruct & 30.43{\tiny \textbf{\textcolor{teal}{(+18.11\%)}}} & 35.20 & \underline{102.30}   & \textbf{96.36} \\
\rowcolor[gray]{.95}
PT & Qwen3-8B-Instruct & 10.14  & 12.38  & 66.21  & 67.15 \\
FT$^{\dagger}$~\cite{xu2024androidlab} & Qwen3-8B-Instruct & 19.56{\tiny \textbf{\textcolor{teal}{(+9.41\%)}}} & 25.60 & 38.69 & 65.18 \\
FT (ours) & Qwen3-8B-Instruct & 26.81{\tiny \textbf{\textcolor{teal}{(+16.66\%)}}} & 31.09 & 72.16 & 69.85 \\
\rowcolor{c2}
RL (ours) & Qwen3-8B-Instruct & \textbf{36.23}{\tiny \textbf{\textcolor{teal}{(+26.08\%)}}} & \textbf{41.96} & 88.04 & 94.49 \\
\rowcolor[gray]{.95}
PT & Qwen3-32B-Instruct & 18.12  & 21.80  & 91.99 & 87.57  \\
FT$^{\dagger}$~\cite{xu2024androidlab} & Qwen3-32B-Instruct & 22.46{\tiny \textbf{\textcolor{teal}{(+4.34\%)}}} & 28.20 & 39.28 & 65.50 \\
FT (ours) & Qwen3-32B-Instruct & 28.98{\tiny \textbf{\textcolor{teal}{(+10.86\%)}}} & 35.92  & 97.79  &  97.33 \\
\rowcolor{c2}
RL (ours) & Qwen3-32B-Instruct & \underline{34.78}{\tiny \textbf{\textcolor{teal}{(+16.66\%)}}} & 40.26  & 89.47  & 93.67  \\
 \bottomrule
\end{tabular}
\begin{tablenotes}
\item [$^{*}$] LLaMA3.1 models only natively support tool calling w/o reasoning.
\item[$^{\dagger}$] The Android Instruct dataset~\cite{xu2024androidlab} is used for fine-tuning where self-verification is not performed.
\item[$^{\ddagger}$] The official results are cited here for comparison.
\end{tablenotes}
\end{threeparttable}
}}}
\end{table}

As shown in Table~\ref{tab:androidlab},
our SmartSnap enables the LLM-driven agents to master interaction with the Android environment during RL the stage,
boosting the task success rate (SR) substantially.
Specifically,
SmartSnap (Qwen3-8B-Instruct) and SmartSnap (Qwen3-32B-Instruct) respectively achieve on-par performance with DeepSeek-V3.1 and Qwen3-235B-A22B,
showcasing quite promising results with regard to their model scales.
These two models support both instruct (i.e., empty thinking \texttt{<think>}\textbackslash n\textbackslash n\texttt{</think>}) and reasoning modes during inference,
but we use the default instruct mode due to the limit of multi-turn context in the present study.
All the models under investigation achieve performance gains over 16\% across model families and sizes,
confirming the generalization of SmartSnap.
It is noted that for LLaMA3.1-8B-Instruct,
its function call mode does not natively support \texttt{ReAct}~\cite{yao2023reactsynergizingreasoningacting} where the reasoning process and the tool calling are interleaved during the rollout progress.
Therefore, its \texttt{act-only} paradigm restricts the explicit semantic representation of state analyses (reflection) and sub-goal decomposition (planning) and thereafter leads to lower upper limit.
Such reasoning-free agent might mechanically repeat its sub-optimal or even incorrect steps,
leading to ineffective acquisition of skill-sets.
This is also confirmed by its comparatively lower reversed redundancy ratio (RRR) where redundant steps are more likely to occur.

Furthermore,
with respect to the performance comparison of SFT,
we find that our Self-Verifying mode achieves much higher task SR than the vanilla Android Instruct dataset~\cite{xu2024androidlab}.
Note that we adopt exactly the same 726 tasks without performing data augmentation for task distribution expansion,
ensuring fair comparison.
Our SFT allows the agent to not only knows how to perform the task but also, more importantly,
understands which kinds of evidences are required and appropriate for curation.
Such self-verifying nature is quite indispensable to the generalization of agent as the guidance on the preparation of key evidences implicitly cultivates the problem decomposition capabilities.
Each evidence can be viewed as a milestone of each sub-task along the path towards completion and thereafter such a curation process contributes to a superior SR.
In contrast,
the vanilla fine-tuning only causes parameterized memorization of solutions. 
It leverages LLMs of various basic capacity (e.g., knowledge QA, reasoning and agentic performance) towards a similar level of 22\% SR.
This means that the mechanic imitation of expert demonstration (e.g., trajectories from a stronger LLM) without evidence-oriented self-verification cannot bring each model to its extreme.

\begin{table}[htbp]
\centering
\caption{Results of success rates per category on AndroidLab with XML mode.
PT, FT, and RL stand for prompting, fine-tuning, and reinforcement learning, respectively.
For all tasks, a higher value means better performance.
}
\label{tab:androidlabcategory}
\resizebox{0.98\linewidth}{!}{
\setlength{\tabcolsep}{1mm}{
\fontsize{9pt}{10pt}\selectfont{
\begin{threeparttable}
\begin{tabular}{lllllllllll}
\toprule
\multirow{2}{*}{Type} & \multirow{2}{*}{Model} & \multicolumn{9}{c}{Success Rate per App Category} \\
 &  & Calendar & Zoom & Bluecoins & PiMusic & Maps.me & Contacts & Cantook & Clock & Setting \\  \midrule
\rowcolor[gray]{.95}
PT & LLaMA3.1-8B-Instruct & 0.00 & 0.00 & 0.00 & 0.00 & 0.00 & 13.33 & 28.57 & 15.38 & 8.33  \\
FT$^{\dagger}$~\cite{xu2024androidlab} & LLaMA3.1-8B-Instruct & 7.14 & 40.00 & 40.00 & 0.00 & 7.14 & 33.33 & 41.66 & 15.38 & 18.18  \\
FT (ours) & LLaMA3.1-8B-Instruct & 7.14 & 20.00 & 26.66 & 8.33 & 0.00 & 60.00 & 33.33 & 14.81 & 39.13 \\
\rowcolor{c2}
RL (ours) & LLaMA3.1-8B-Instruct & 7.14 & 60.00 & 20.00  & 8.33 & 0.00  & 53.33 & 50.00 & 37.03 & 47.82 \\
\rowcolor[gray]{.95}
PT & Qwen2.5-7B-Instruct & 0.00 & 40.00 & 10.00 & 10.00 & 0.00 & 20.00 & 16.66 & 18.51 & 14.28  \\
FT$^{\dagger}$~\cite{xu2024androidlab} & Qwen2.5-7B-Instruct & 14.28 & 40.00 & 13.33 & 8.33 & 0.00 & 46.66  & 33.33 & 18.51 & 21.73 \\
FT (ours) & Qwen2.5-7B-Instruct & 21.42 & 40.00 & 40.00 & 8.33 & 0.00 & 46.66 & 41.66 & 33.33 & 43.47 \\
\rowcolor{c2}
RL (ours) & Qwen2.5-7B-Instruct & 14.28 & 20.00 & 33.33 & 8.33 & 0.00 & 60.00 & 50.00 & 37.03 & 34.78   \\
\rowcolor[gray]{.95}
PT & Qwen3-8B-Instruct & 7.14 & 66.66  & 0.00 & 8.33 & 0.00 & 15.38 & 11.11 & 4.00 & 15.00 \\
FT$^{\dagger}$~\cite{xu2024androidlab} & Qwen3-8B-Instruct & 7.14 & 40.00 & 6.66 & 8.33 & 0.00 & 33.33 & 50.00 & 18.51 & 26.08 \\
FT (ours) & Qwen3-8B-Instruct & 21.42 & 60.00 & 53.33 & 8.33 & 0.00 & 16.66 & 45.45 & 40.74 & 41.66 \\
\rowcolor{c2}
RL (ours) & Qwen3-8B-Instruct & 14.28 & 60.00 & 33.33 & 33.33 & 13.33 & 53.33 & 50.00 & 40.74 & 39.13  \\
\rowcolor[gray]{.95}
PT & Qwen3-32B-Instruct & 0.00 & 80.00 & 13.33 & 0.00 & 0.00 & 33.33 & 33.33 &  22.22 & 17.39 \\
FT$^{\dagger}$~\cite{xu2024androidlab} & Qwen3-32B-Instruct & 14.28 & 60.00 & 26.66 & 0.00 & 0.00 & 40.00 & 50.00 & 25.92 & 13.04 \\
FT (ours) & Qwen3-32B-Instruct & 7.14   &  40.00  &  13.33  & 8.33   &  0.00  &  53.33  &  50.00  &  40.74  &  39.13  \\
\rowcolor{c2}
RL (ours) & Qwen3-32B-Instruct & 7.14 & 80.00 & 40.00 & 25.00 & 0.00 & 60.00  & 50.00  & 38.46 & 39.13 \\
 \bottomrule
\end{tabular}
\begin{tablenotes}
\item[$^{\dagger}$] The Android Instruct dataset~\cite{xu2024androidlab} is used for fine-tuning where self-verification is not performed.
\end{tablenotes}
\end{threeparttable}
}}}
\end{table}

Table~\ref{tab:androidlabcategory} provides the detailed success rates per category.
Compared with the naive prompting and the vanilla instruction fine-tuning,
our SFT and RL achieve performance gains consistently on most of the app categories except Maps.me.
Specifically,
we find that the relative performance gains are closely associated with the distributions of training tasks.
As shown in Table~\ref{tab:androidlabdistribution},
the tasks from App \texttt{Setting} are dominating and thereafter steady improvement is observed on this domain for both SFT and RL stages.
This highlights that for further studies on mobile agents,
the preparation of training data should emphasize:
1) the balance between tasks of different apps,
and 2) the diversity of tasks within each app.
Moreover,
we observe that the difficulty of tasks also plays a critical role in RL effectiveness.
The almost zero variation of performance (both SFT and RL) on App \texttt{Maps.me},
regardless of model families and scales,
implies that there exists a huge knowledge gap between the existing pretrained LLMs and the mobile domains.
Without intensive knowledge injection,
the agent itself cannot explore effectively to handle tasks such as finding the shortest path via a specific transportation or locating the destination that meets all target requirements.

\subsection{Qualitative Analysis}

We present a comprehensive analysis and highlight several key findings from our RL dynamics.
\begin{thmbox}
\paragraph{Takeaways}
\begin{itemize}
    \item The agents manage to submit only relevant and persuasive evidences with the number of evidences converging towards 1.5 on average.
    \item The agents learn to complete tasks and submit evidences in a more compact and efficient interaction manner without mindless trials and errors.
    \item The RL itself increases the training reward consistently with a decreasing intra-group variance, showcasing that the agents gradually overfit the few available training tasks.
    \item The fluctuation of performance on complex domains like Calendar, Maps.me, and Zoom suggests that agents struggle at deriving robust strategies without knowledge supplement. 
\end{itemize}
\end{thmbox}

\begin{figure}[htbp]
\begin{center}
\begin{subfigure}{.33\textwidth}
  \centering
  \includegraphics[width=\textwidth]{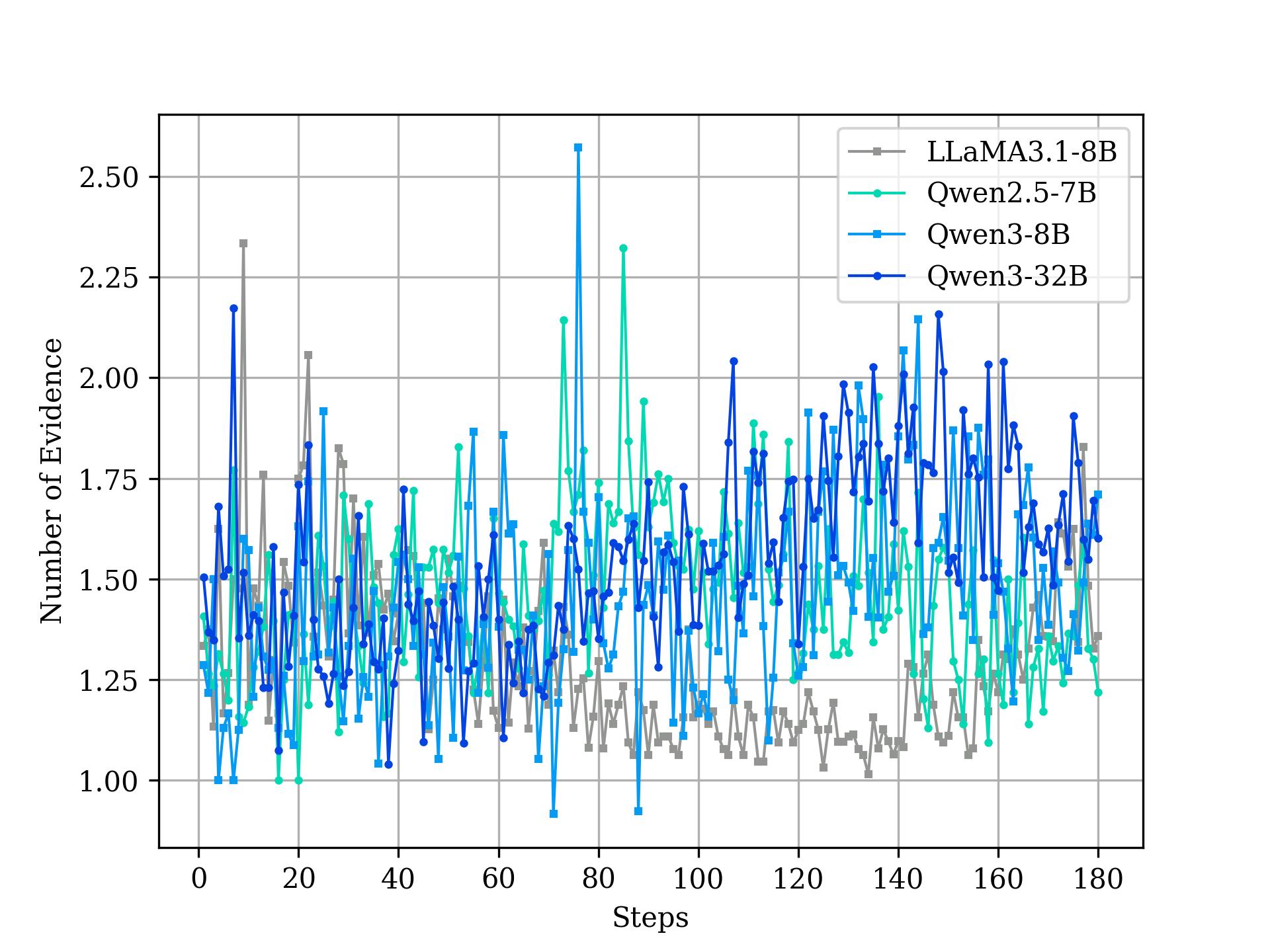}
  \caption{Number of Evidence (Training).}
\end{subfigure}
\begin{subfigure}{.33\textwidth}
  \centering
  \includegraphics[width=\textwidth]{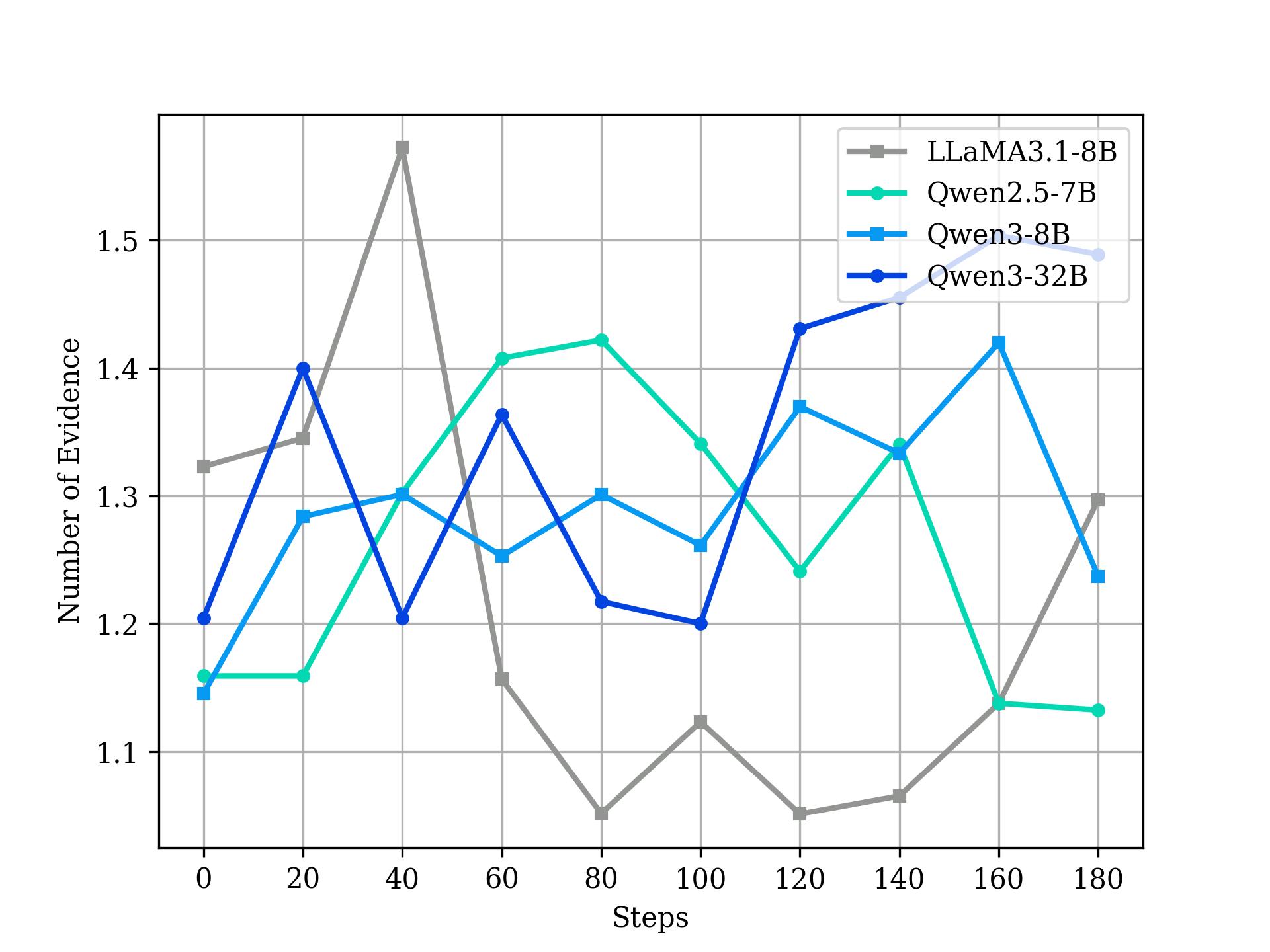}
  \caption{Number of Evidence (Validation).}
\end{subfigure}%
\begin{subfigure}{.33\textwidth}
  \centering
  \includegraphics[width=\textwidth]{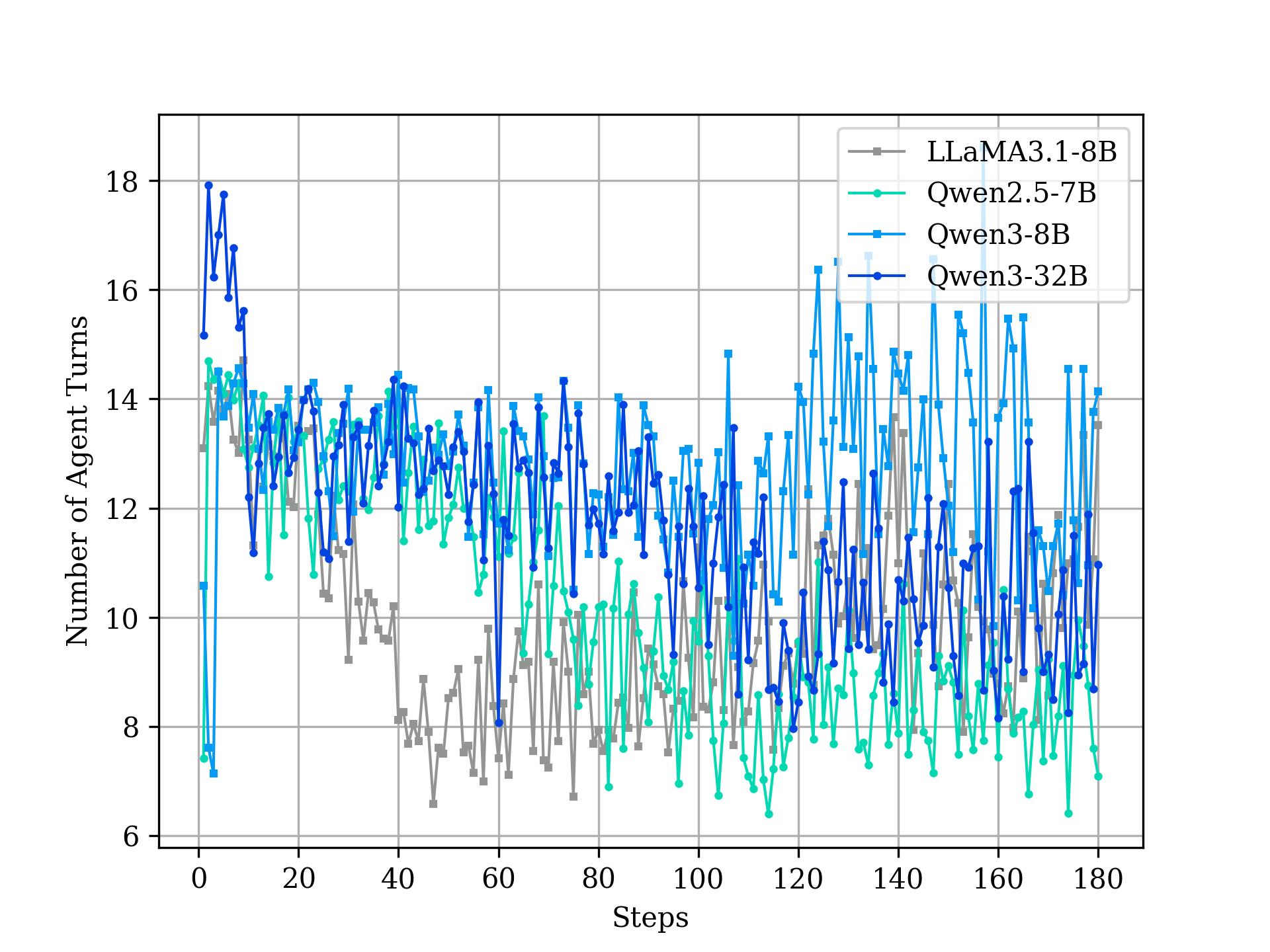}
  \caption{Number of Agent Turns (Training).}
\end{subfigure}%
\\
\begin{subfigure}{.33\textwidth}
  \centering
  \includegraphics[width=\textwidth]{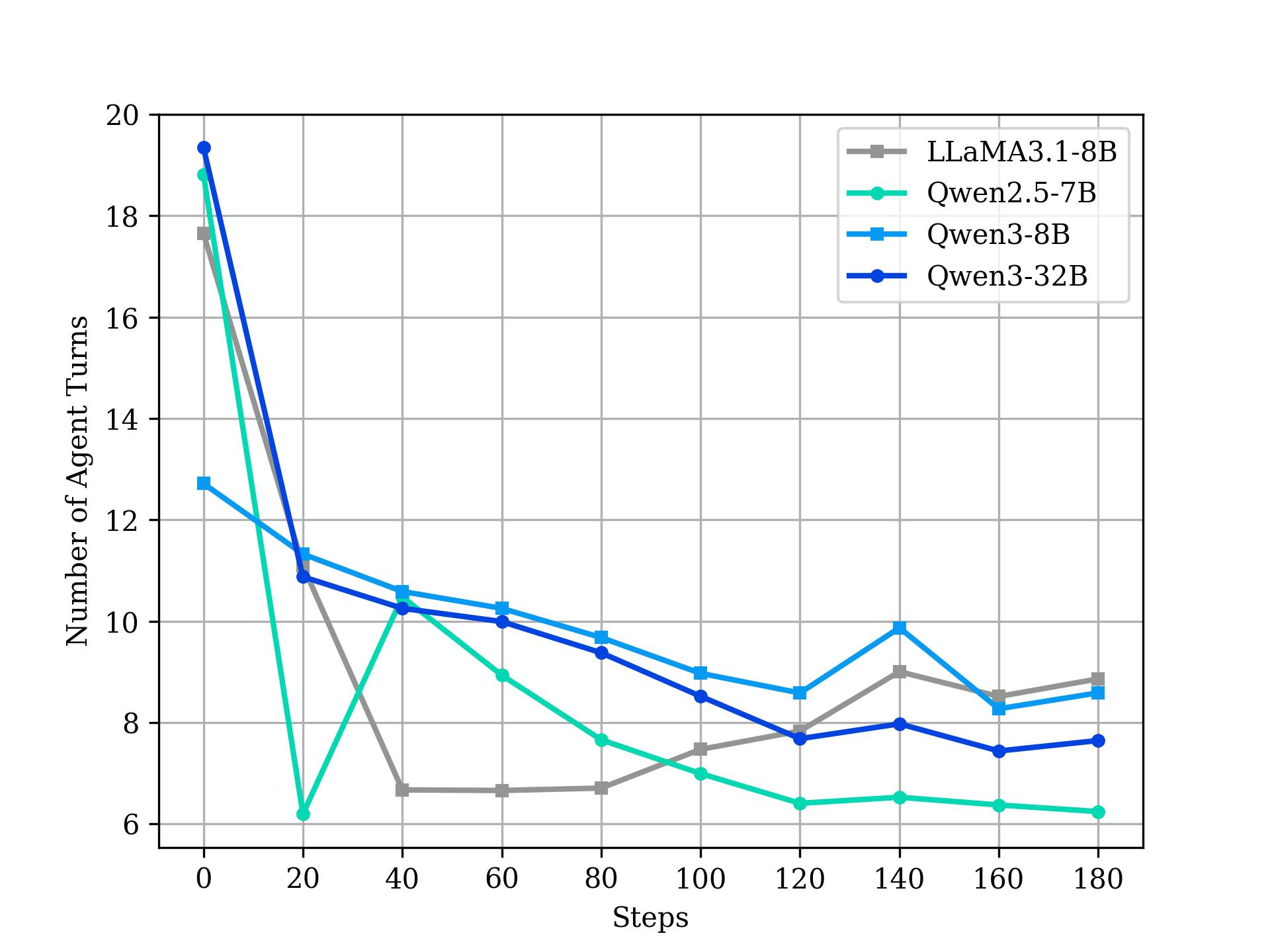}
  \caption{Number of Agent Turns (Validation).}
\end{subfigure}
\begin{subfigure}{.33\textwidth}
  \centering
  \includegraphics[width=\textwidth]{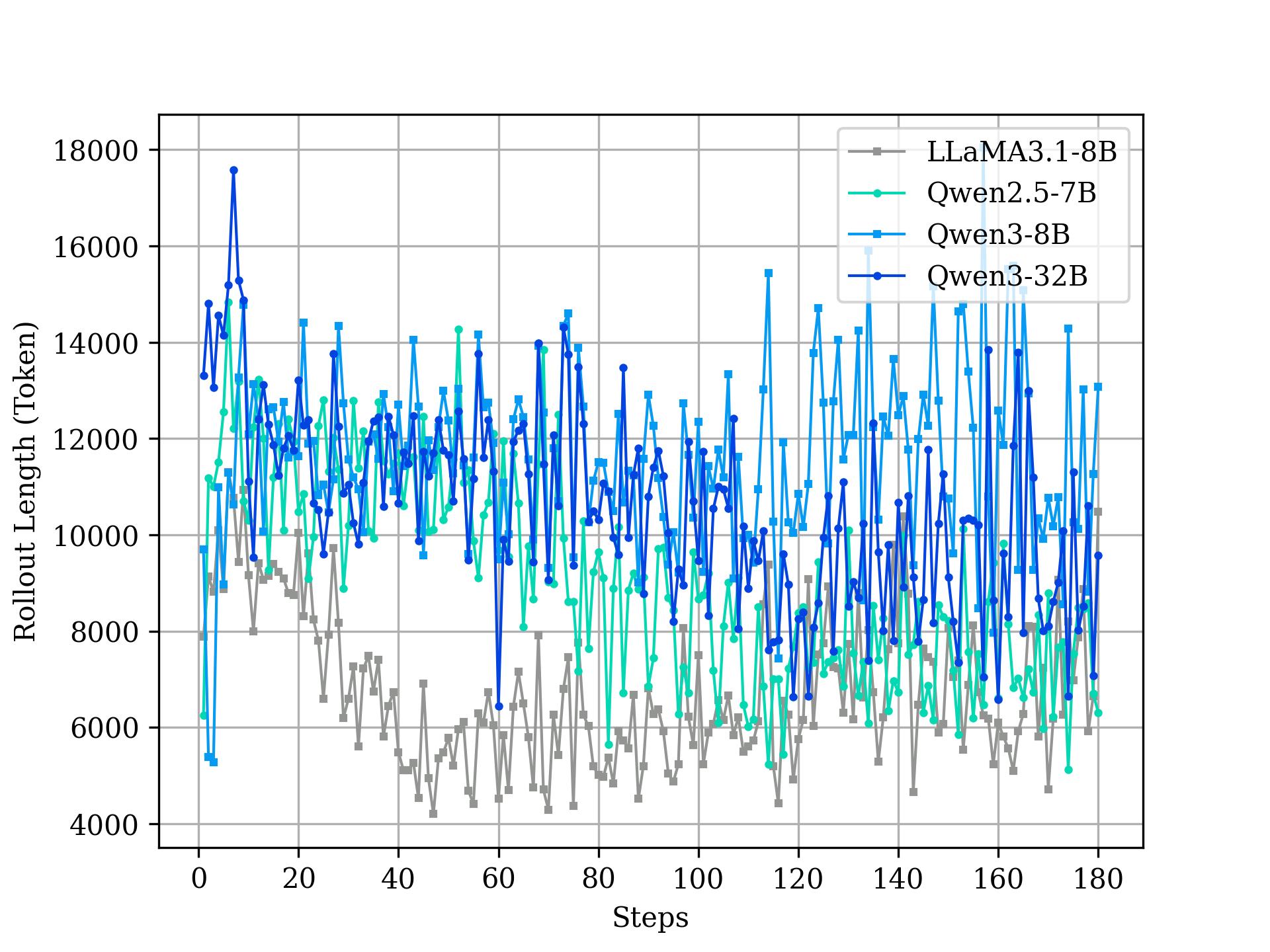}
  \caption{Response Length (Training).}
\end{subfigure}
\begin{subfigure}{.33\textwidth}
  \centering
  \includegraphics[width=\textwidth]{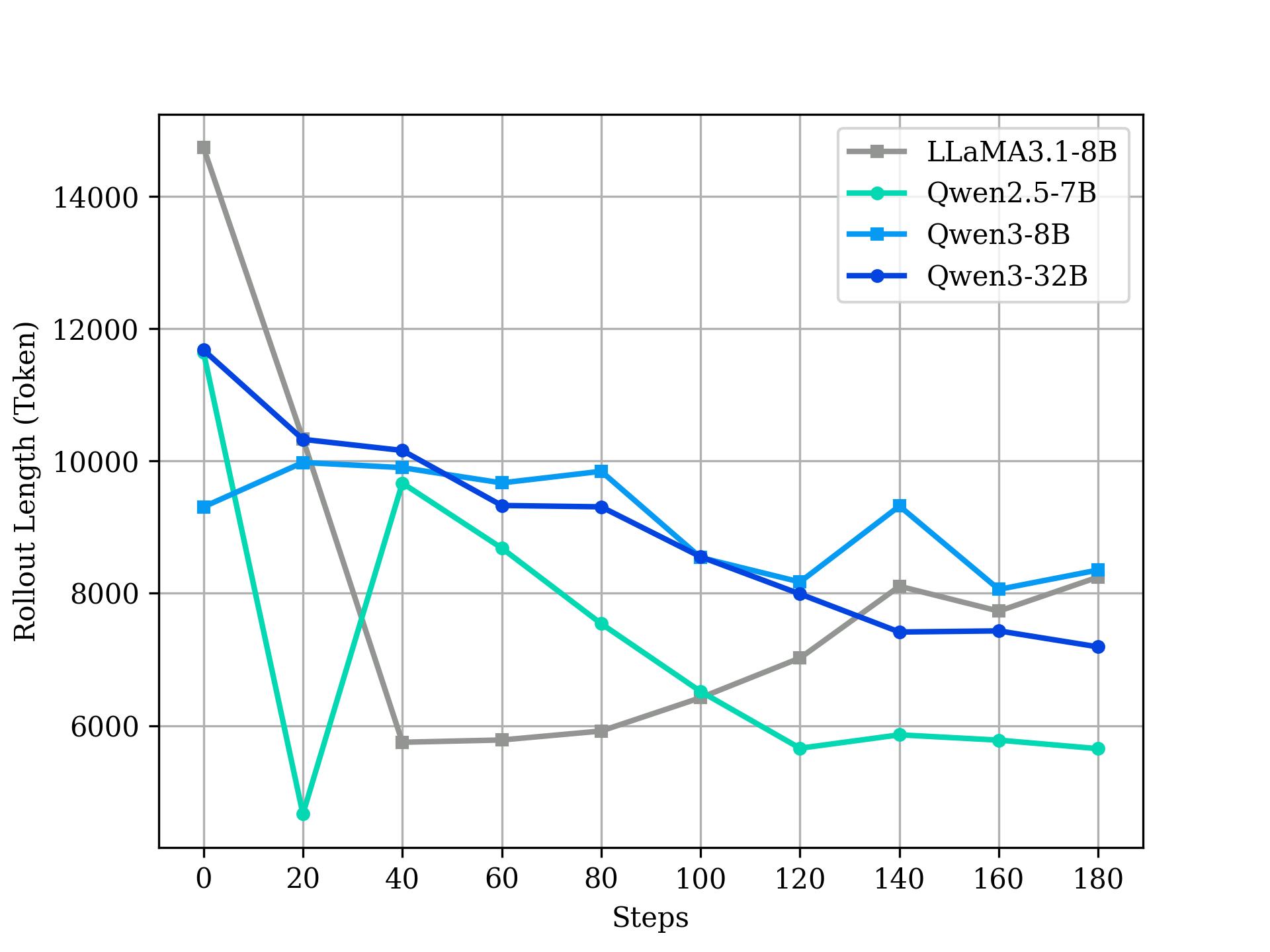}
  \caption{Response Length (Validation).}
\end{subfigure}
\end{center}
\caption{Agent behavior evolution (RL dynamics).}
\label{fig:training_dynamics}
\end{figure}

\begin{figure}[htbp]
\begin{center}
\begin{subfigure}{.33\textwidth}
  \centering
  \includegraphics[width=\textwidth]{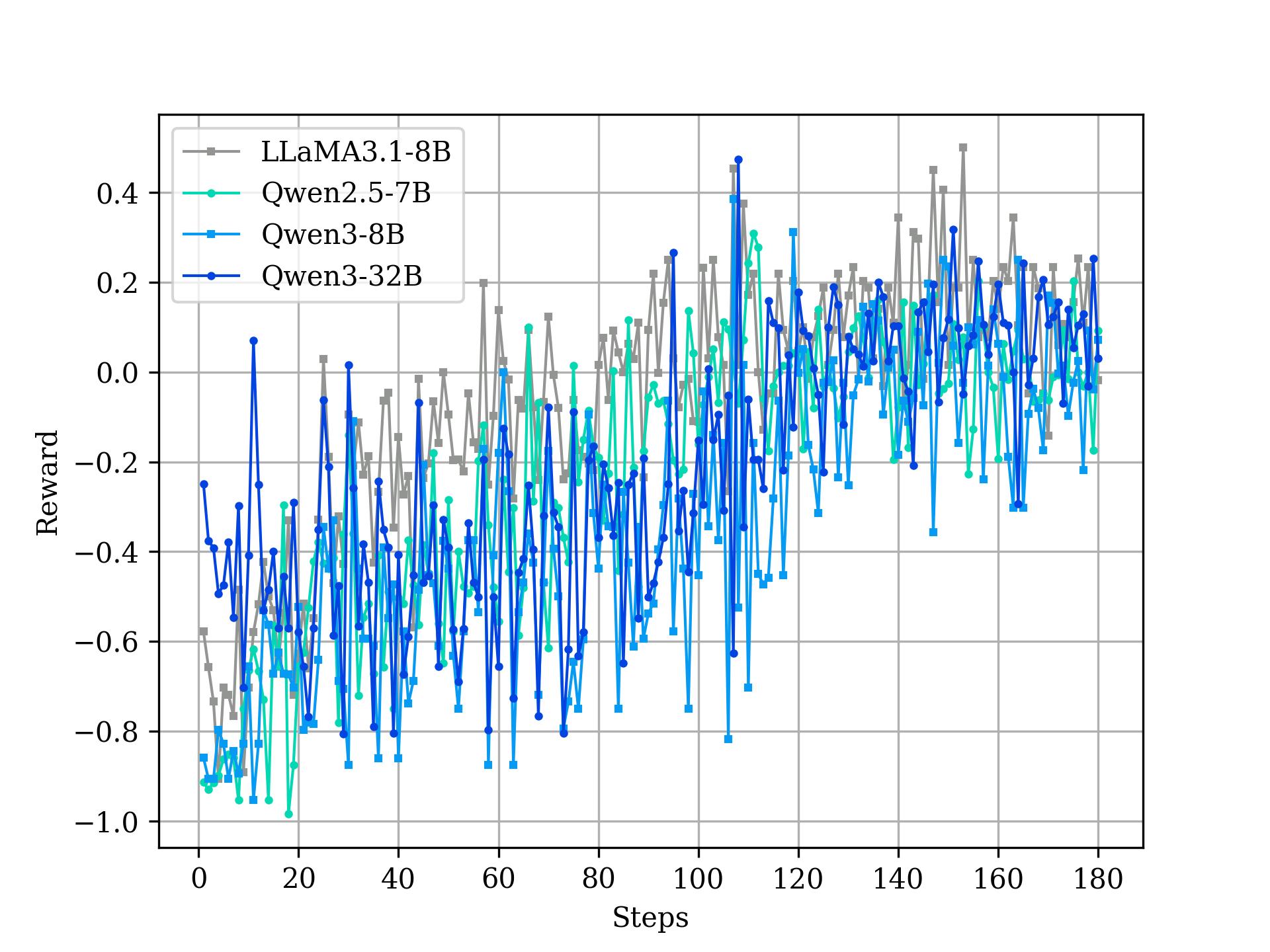}
  \caption{Training Reward (mean).}
\end{subfigure}
\hspace{1em}
\begin{subfigure}{.33\textwidth}
  \centering
  \includegraphics[width=\textwidth]{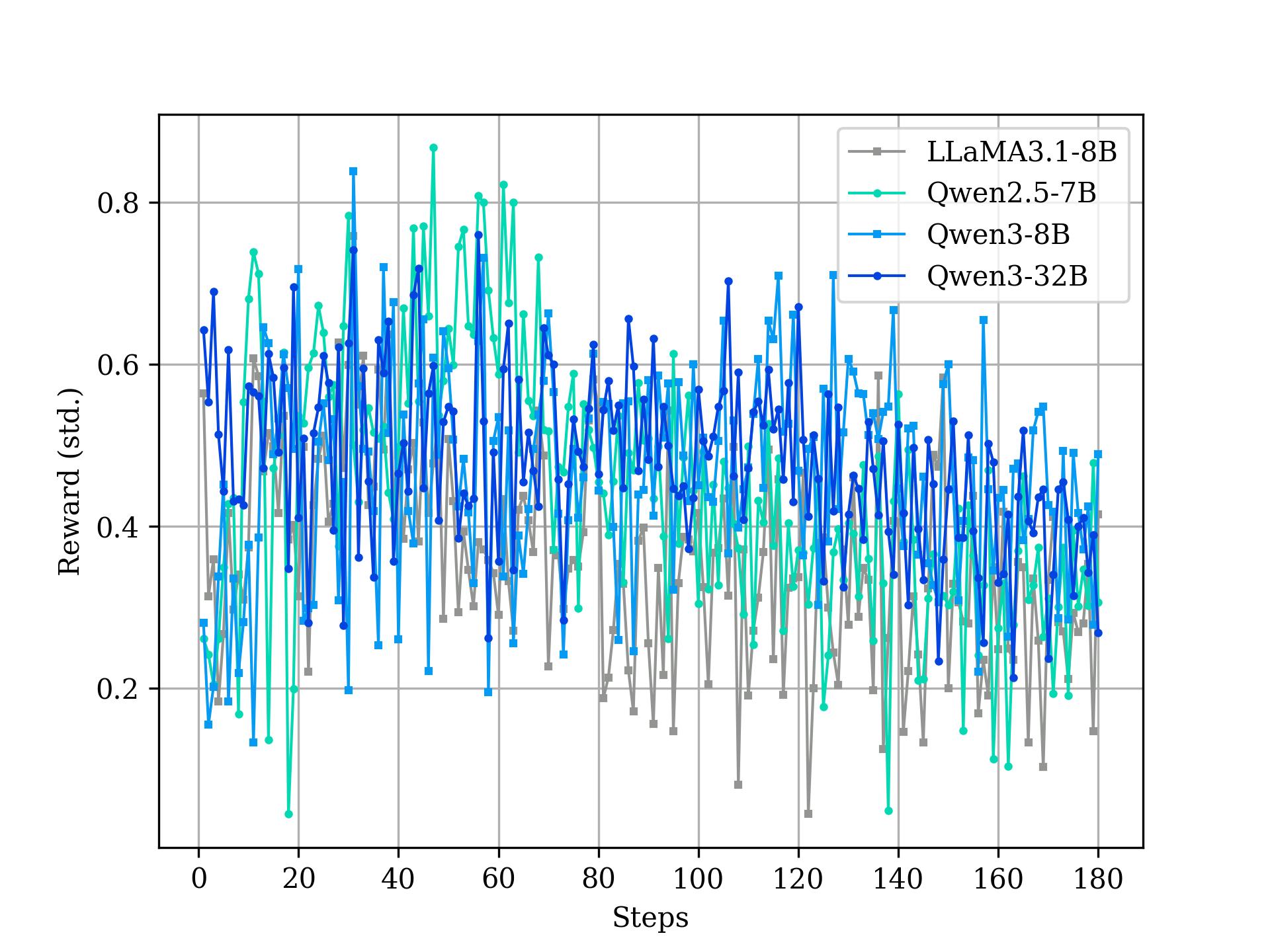}
  \caption{Training Reward (std.).}
\end{subfigure}
\\
\begin{subfigure}{.33\textwidth}
  \centering
  \includegraphics[width=\textwidth]{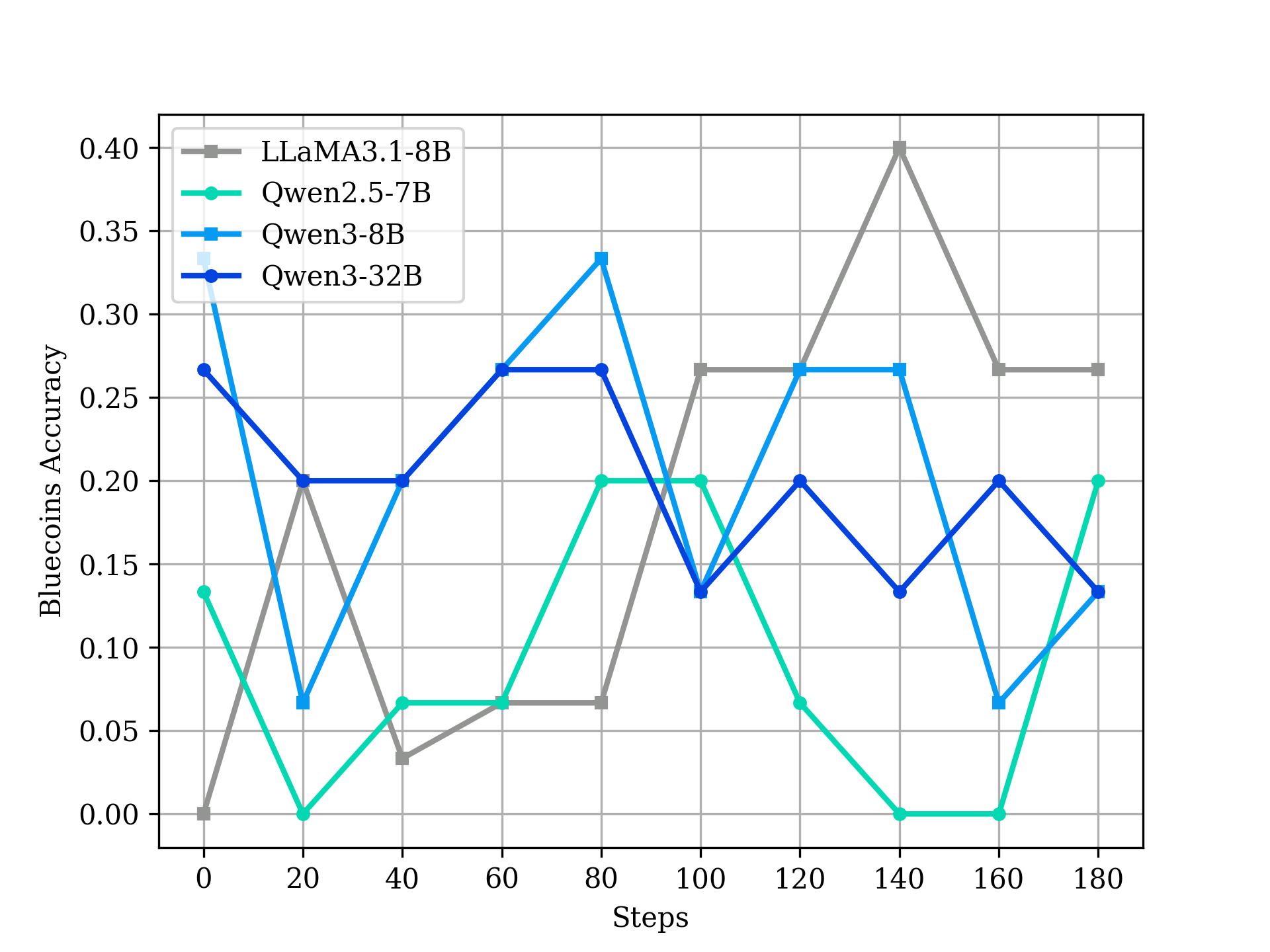}
  \caption{Validation on Bluecoins.}
\end{subfigure}%
\begin{subfigure}{.33\textwidth}
  \centering
  \includegraphics[width=\textwidth]{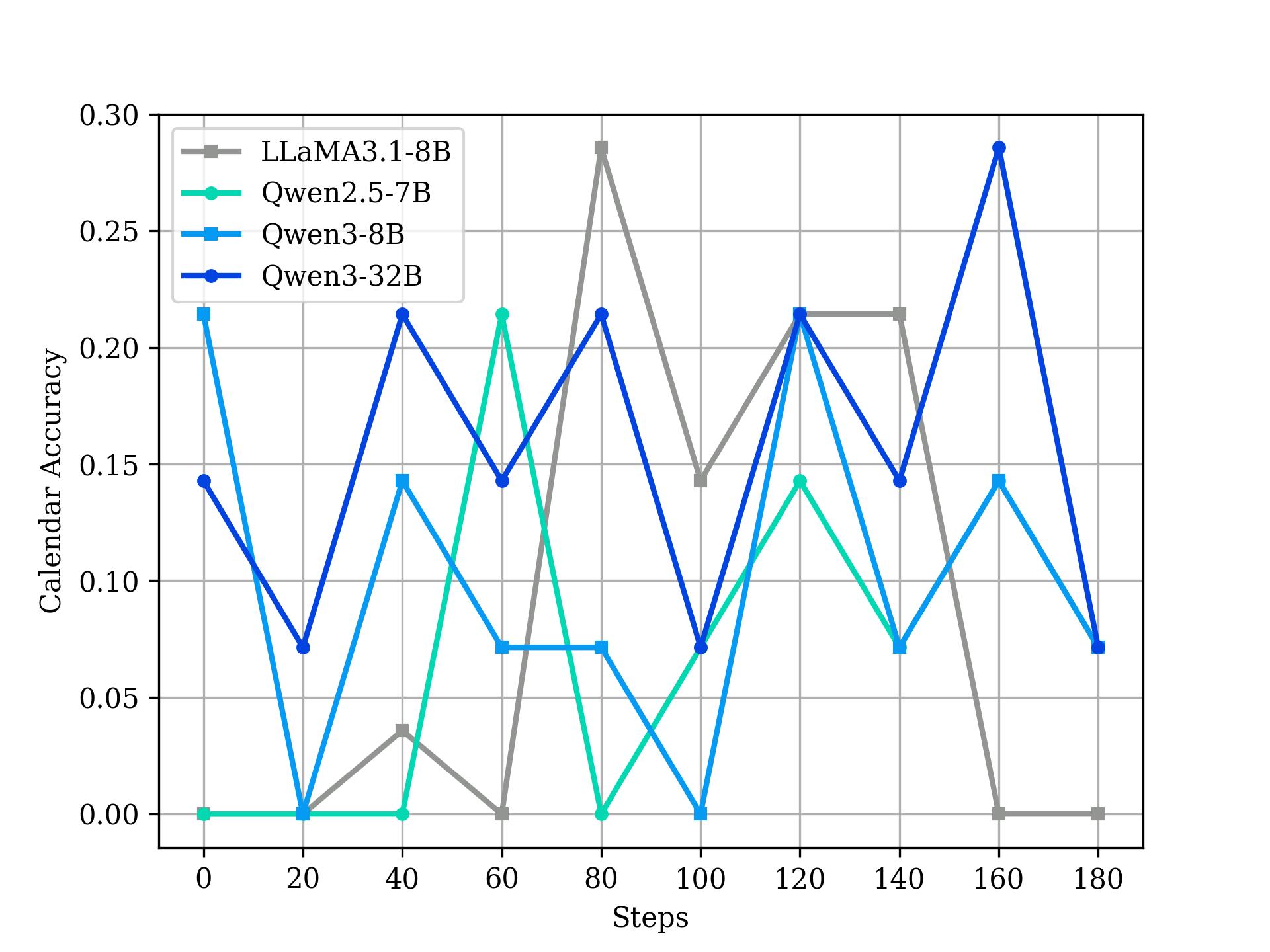}
  \caption{Validation on Calendar.}
\end{subfigure}%
\begin{subfigure}{.33\textwidth}
  \centering
  \includegraphics[width=\textwidth]{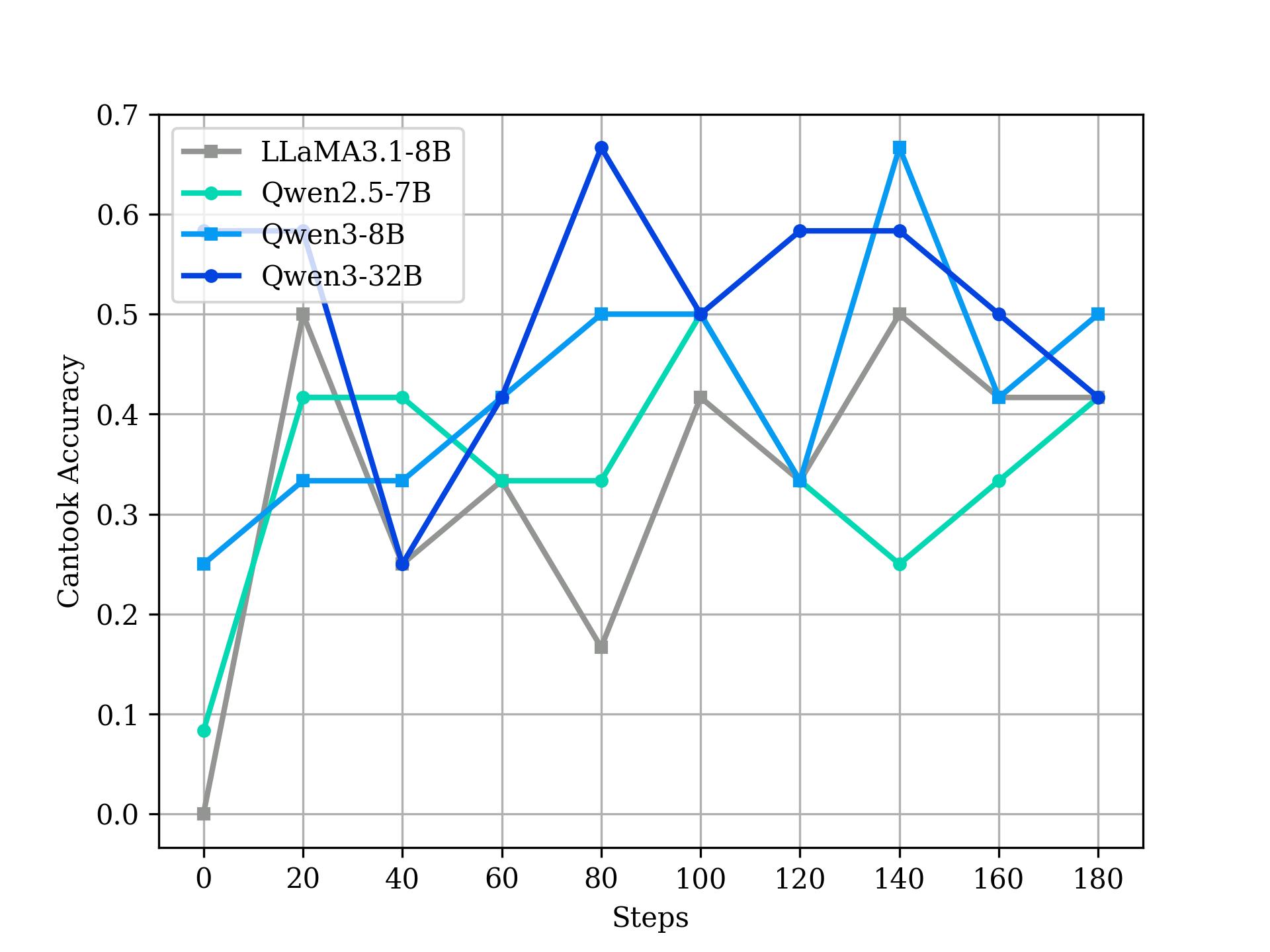}
  \caption{Validation on Cantook.}
\end{subfigure}%
\\
\begin{subfigure}{.33\textwidth}
  \centering
  \includegraphics[width=\textwidth]{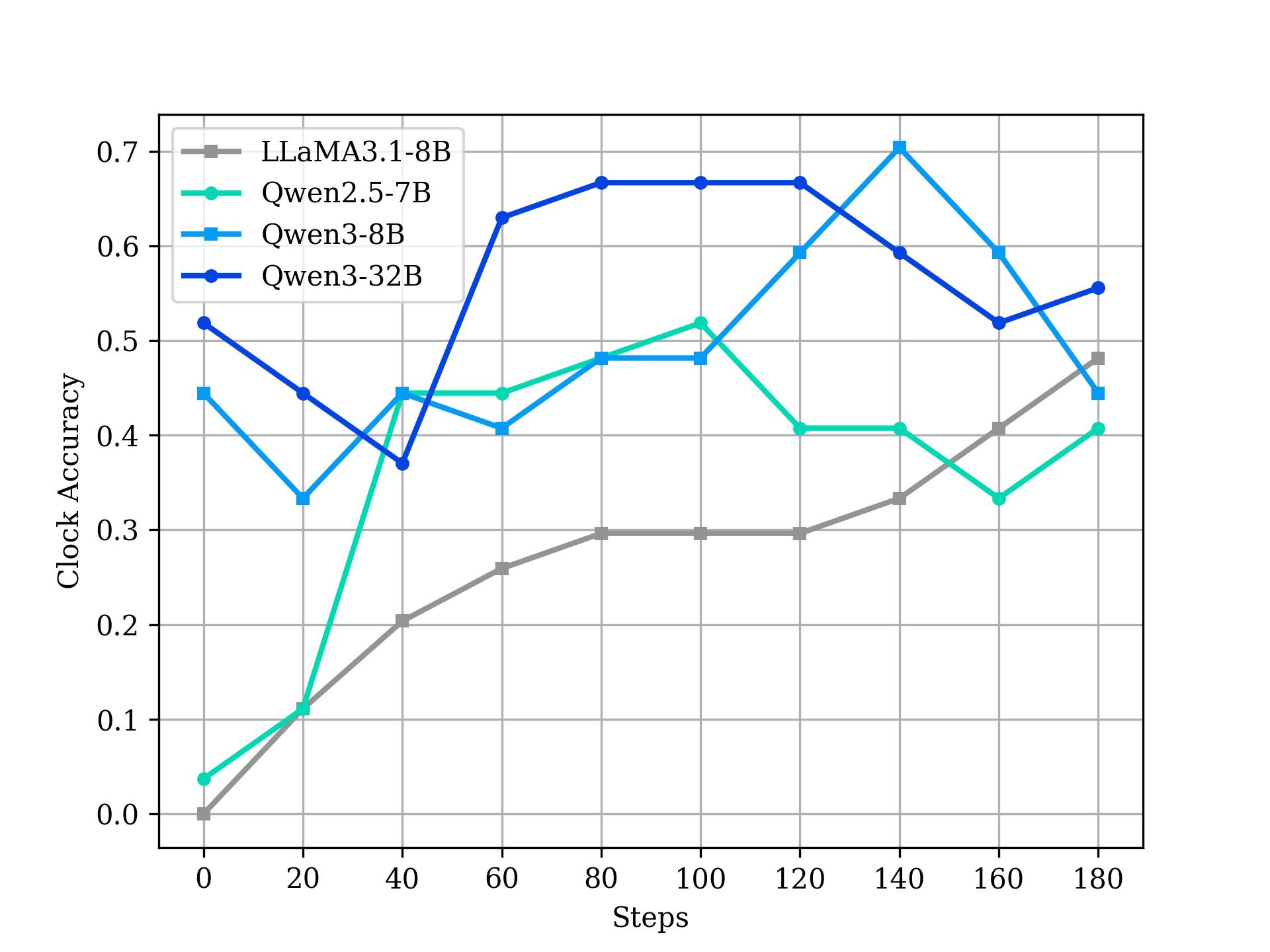}
  \caption{Validation on Clock.}
\end{subfigure}%
\begin{subfigure}{.33\textwidth}
  \centering
  \includegraphics[width=\textwidth]{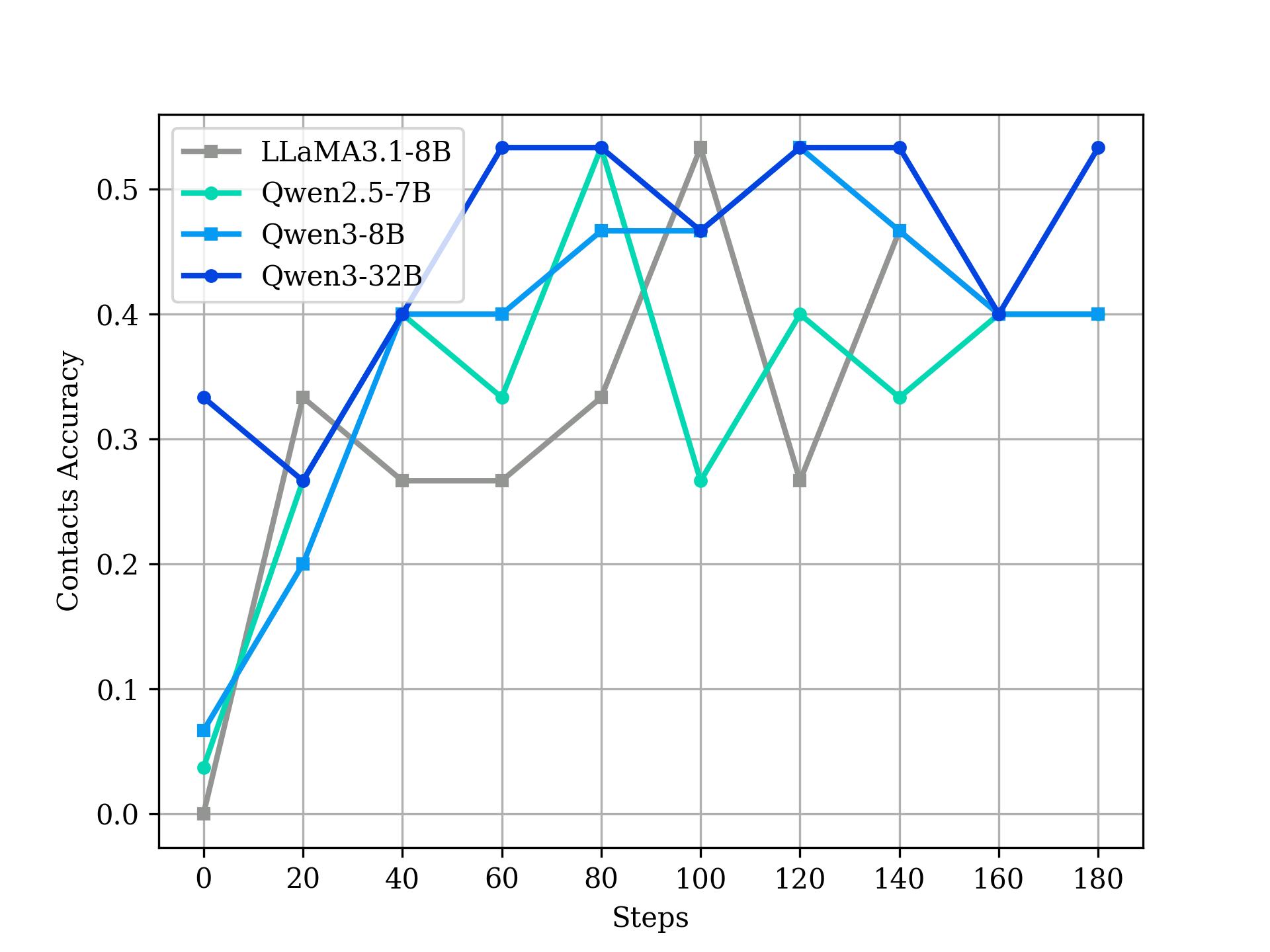}
  \caption{Validation on Contacts.}
\end{subfigure}%
\begin{subfigure}{.33\textwidth}
  \centering
  \includegraphics[width=\textwidth]{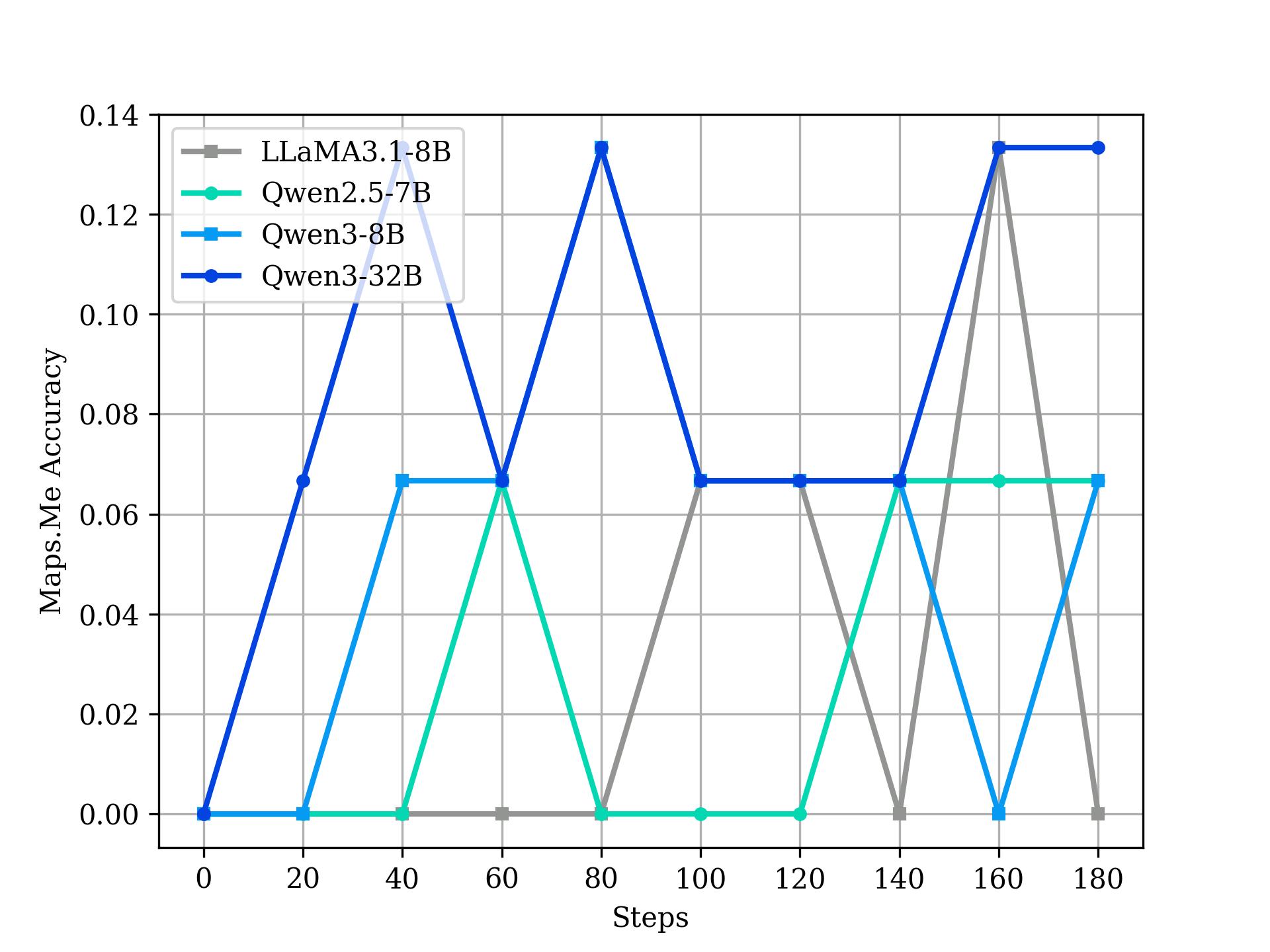}
  \caption{Validation on Maps.me.}
\end{subfigure}%
\\
\begin{subfigure}{.33\textwidth}
  \centering
  \includegraphics[width=\textwidth]{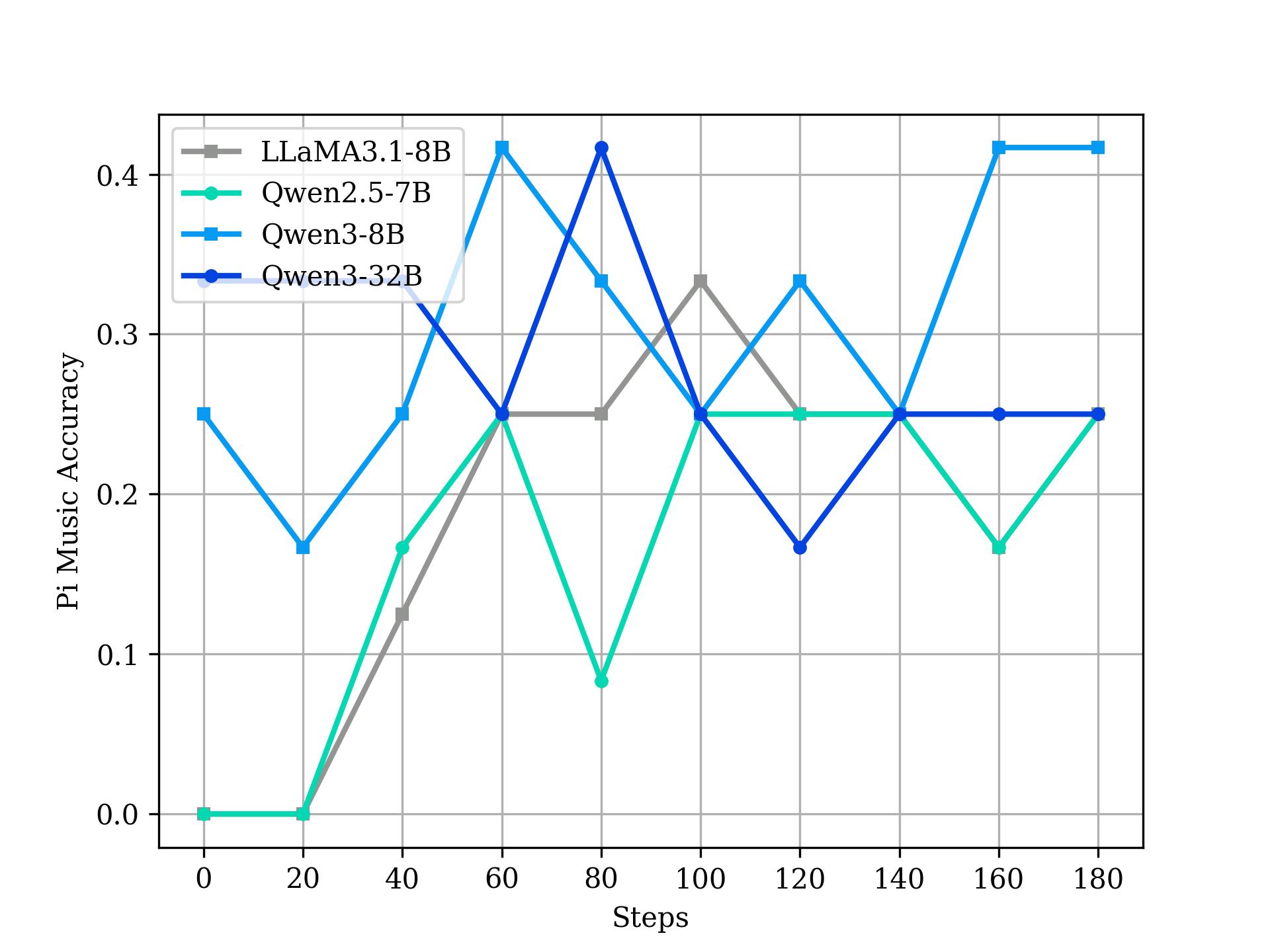}
  \caption{Validation on Pi Music.}
\end{subfigure}%
\begin{subfigure}{.33\textwidth}
  \centering
  \includegraphics[width=\textwidth]{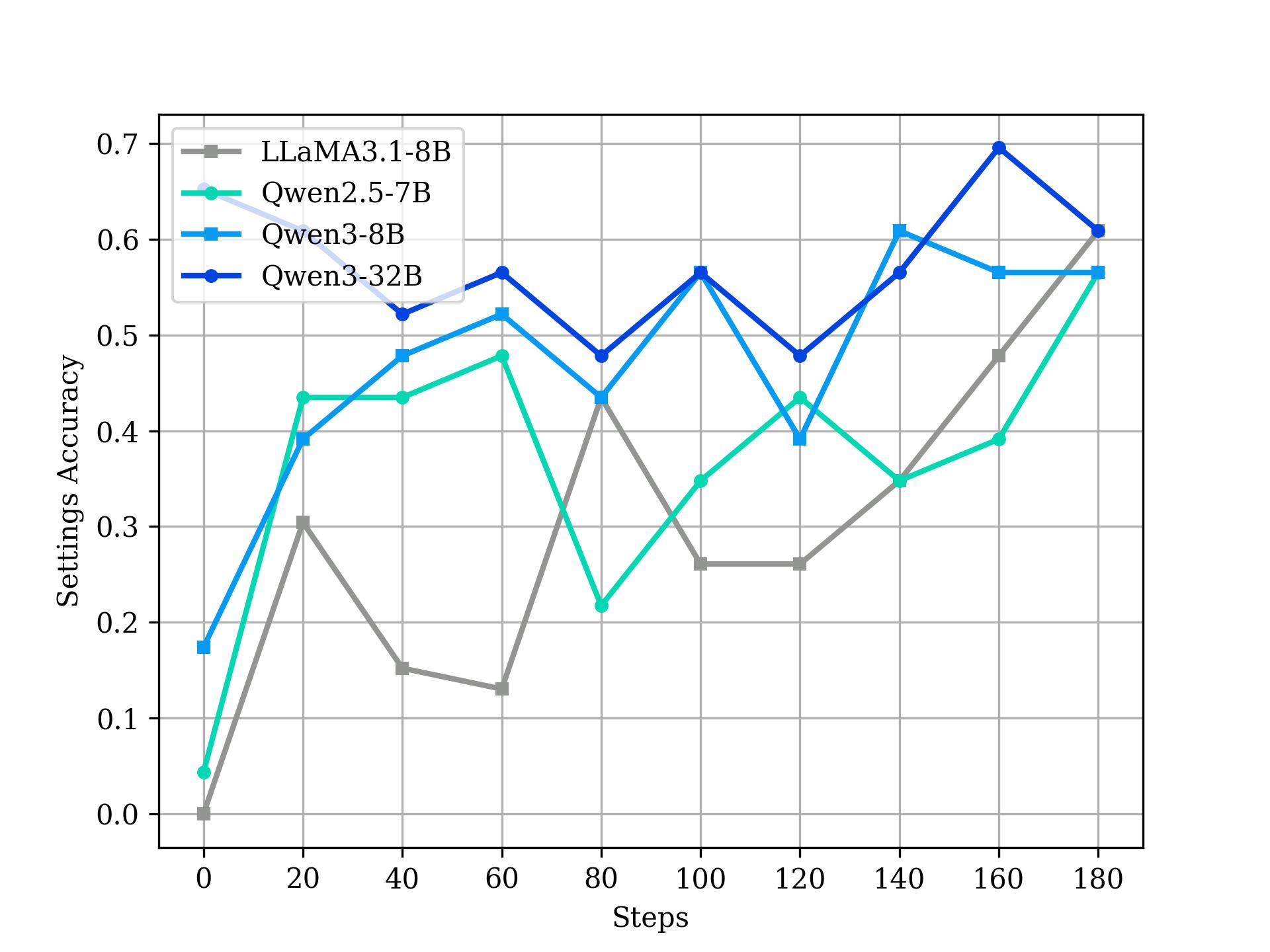}
  \caption{Validation on Settings.}
\end{subfigure}%
\begin{subfigure}{.33\textwidth}
  \centering
  \includegraphics[width=\textwidth]{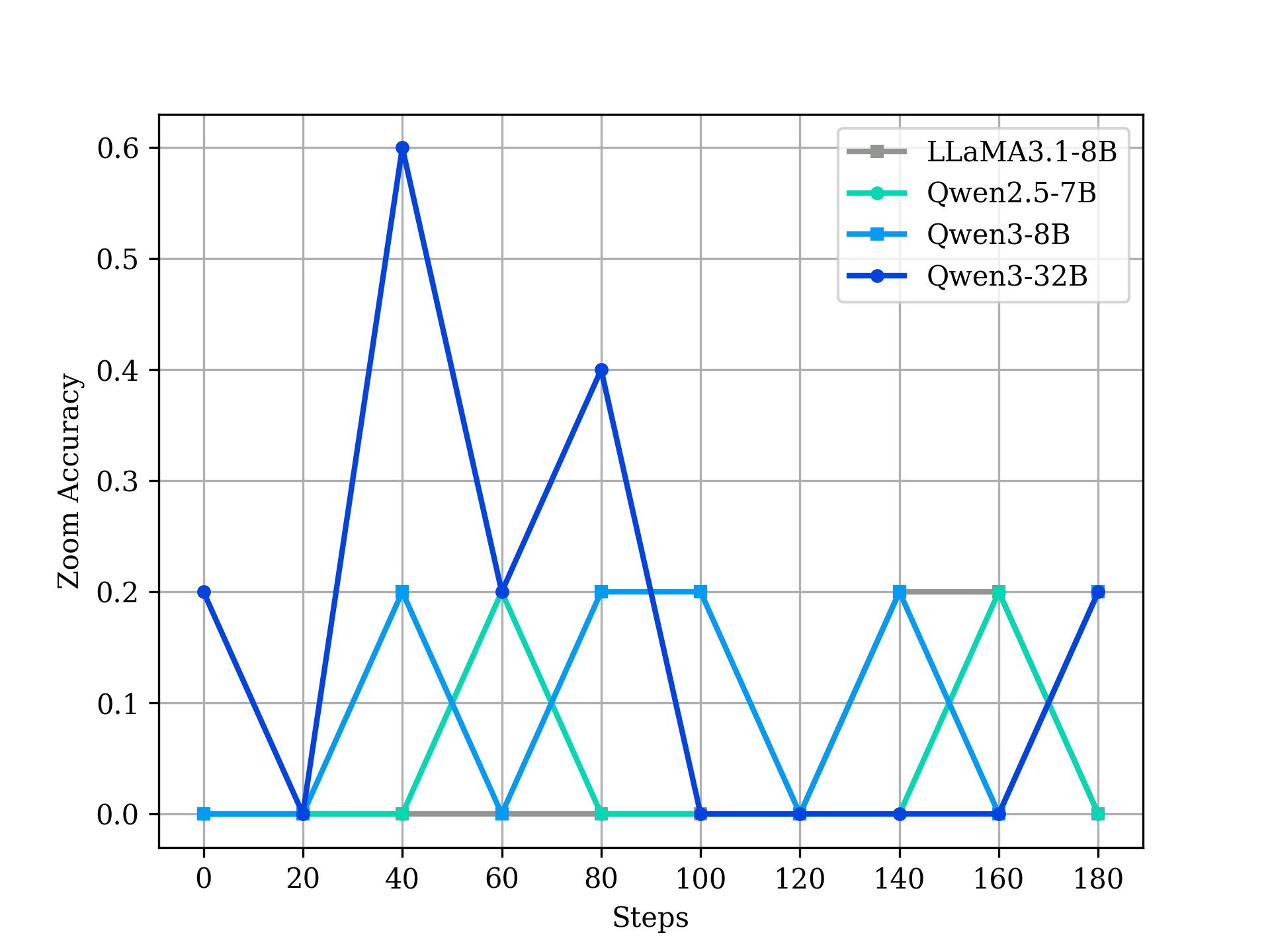}
  \caption{Validation on Zoom.}
\end{subfigure}%
\end{center}
\caption{Agent performance variation (RL dynamics).}
\label{fig:training_dynamics2}
\end{figure}

\paragraph{Behavior Evolution}
To gain deeper insights into the RL process of SmartSnap,
we illustrate the evolution of agent behaviors in Figure~\ref{fig:training_dynamics}.
\begin{itemize}
    \item{Number of Evidence}: The number of evidences is penalized to be smaller than 3 in our intrinsic reward design. It not only prevents potential reward hacking where the overwhelming evidences might cheat the LLM-as-a-Judge in return for higher rewards, but also encourages the agent to create or select only truly relevant evidences for verification. For training, the \texttt{Act-only} LLaMA3.1-8B exhibits a downward-then-upward trend while the \texttt{ReAct} Qwen series demonstrate more complex patterns. We find that the LLaMA3.1-8B tends to submit a few last screenshots as evidences at the early training stages,
    and then proposes to submit only the last one as training progresses.
    Its relatively limited evidence creativity suggests that a brief thinking before acting is beneficial to the maintenance of evidences.
    \item{Number of Agent Turns}:  All models across families and scales undergo a decreasing trend of interaction turns for both training and validation. Given the monotonous increase of training reward, we believe SmartSnap encourages agents to complete tasks in a more efficient manner.
    \item{Response of Length}: In line with the number of agent turns, the response length decreases accordingly. During the RL stage, agents tend to exploit the existing successful patterns and becomes more proficient when they encounter the same tasks again during the second training epoch.
\end{itemize}

\paragraph{Performance Variation}
To investigate the performance variation across training steps,
we provide the both the training reward and the validation accuracy in Figure~\ref{fig:training_dynamics2}.
Note that the accuracy score reported here is delivered by the LLM-as-a-Judge,
which reflects the performance of both task solving and evidence curation capabilities.
\begin{itemize}
    \item{Training Reward}: It is observed that the average training reward consistently improves, demonstrating that the agent indeed learns to complete the tasks and curate valid snapshot evidences.
    The decreasing intra-group standard deviation also suggests that during RL, the agent gradually overfits the few 726 tasks with a group of trajectories either full success or failure. To maintain effective RL signals, we believe the construction of a larger, difficulty-balanced training set is indispensable.
    \item{Validation Accuracy}: The validation accuracy increases on most app domains except Calendar, Maps.me, and Zoom. On these three apps, we find that agents of all model families and scales are struggling with a fluctuating performance close to zero. One reason is that the testing samples from these categories are too scarce to provide smooth estimation of performance. Another underlying reason is that agents fail to develop robust tactics to handle these tasks, revealing a potential knowledge gap of LLMs under mobile scenarios.
    One would have to perform continual pre-training (CPT) to inject domain-specific knowledge, allowing shallow-yet-braod coverage of operation skills~\cite{yue2025doesreinforcementlearningreally,zhang2025interplaypretrainingmidtrainingrl}.
\end{itemize}

\begin{figure}[htbp]
\begin{center}
\includegraphics[width=\linewidth]{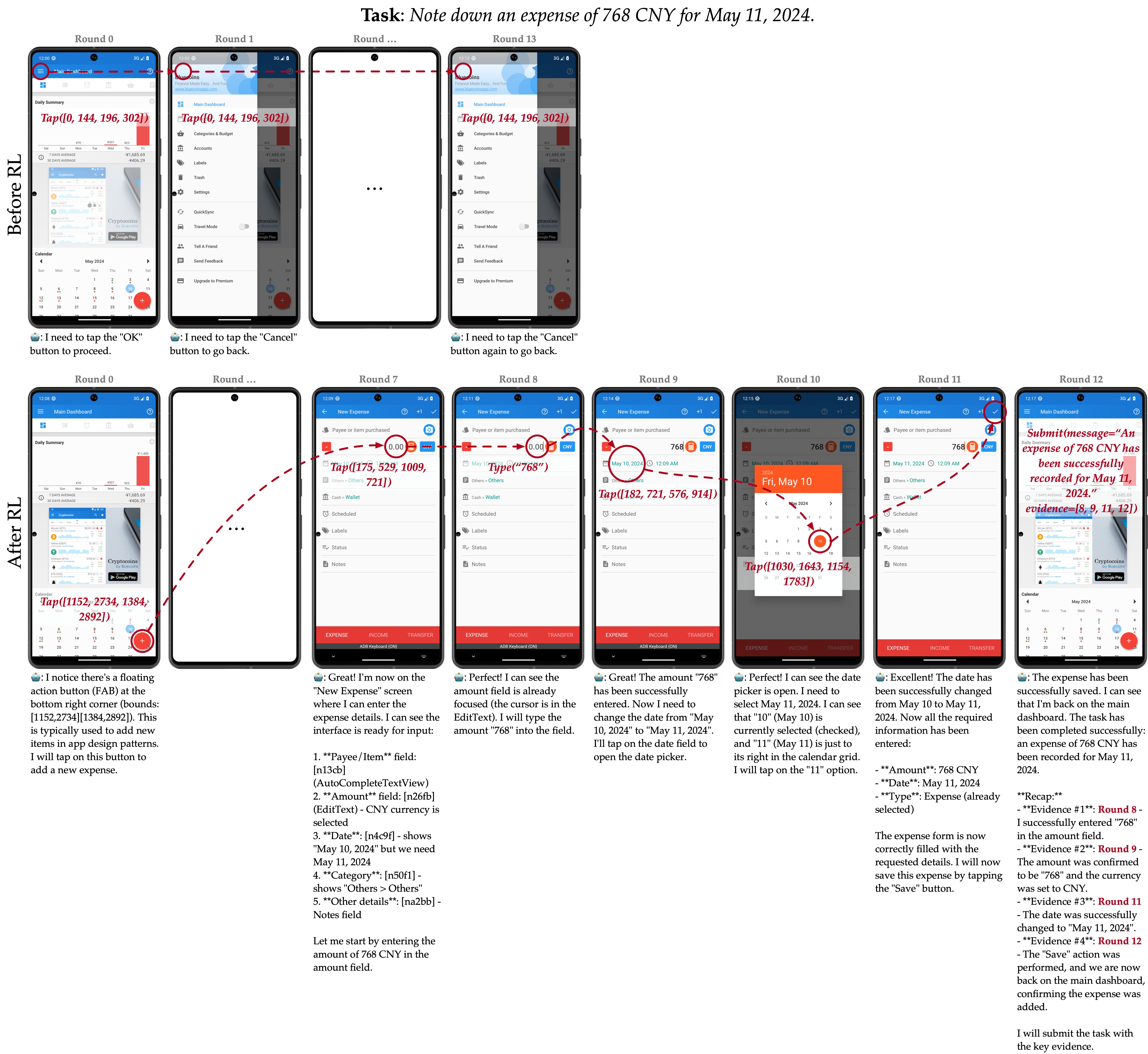}
\end{center}
\caption{
The agent trajectories before and after SmartSnap RL of one testing example of the app Bluecoins (Qwen3-8B).
Initially,
the agent fails to comprehend the instruction properly and consistently taps the top-left \texttt{ok/cancel} button to find the entrance towards creating a new expense note.
After SmartSnap RL,
the agent learns to utilize the floating button for creating new notes with correct in-filling of amount and date.
In addition,
it grasps the evidence-based self-verification during stepwise reasoning,
where snapshots that respectively prove the completion of filling amount, date, and expense type are submitted for verification.
}
\label{fig:app_bluecoin}
\end{figure}

\begin{figure}[htbp]
\begin{center}
\includegraphics[width=\linewidth]{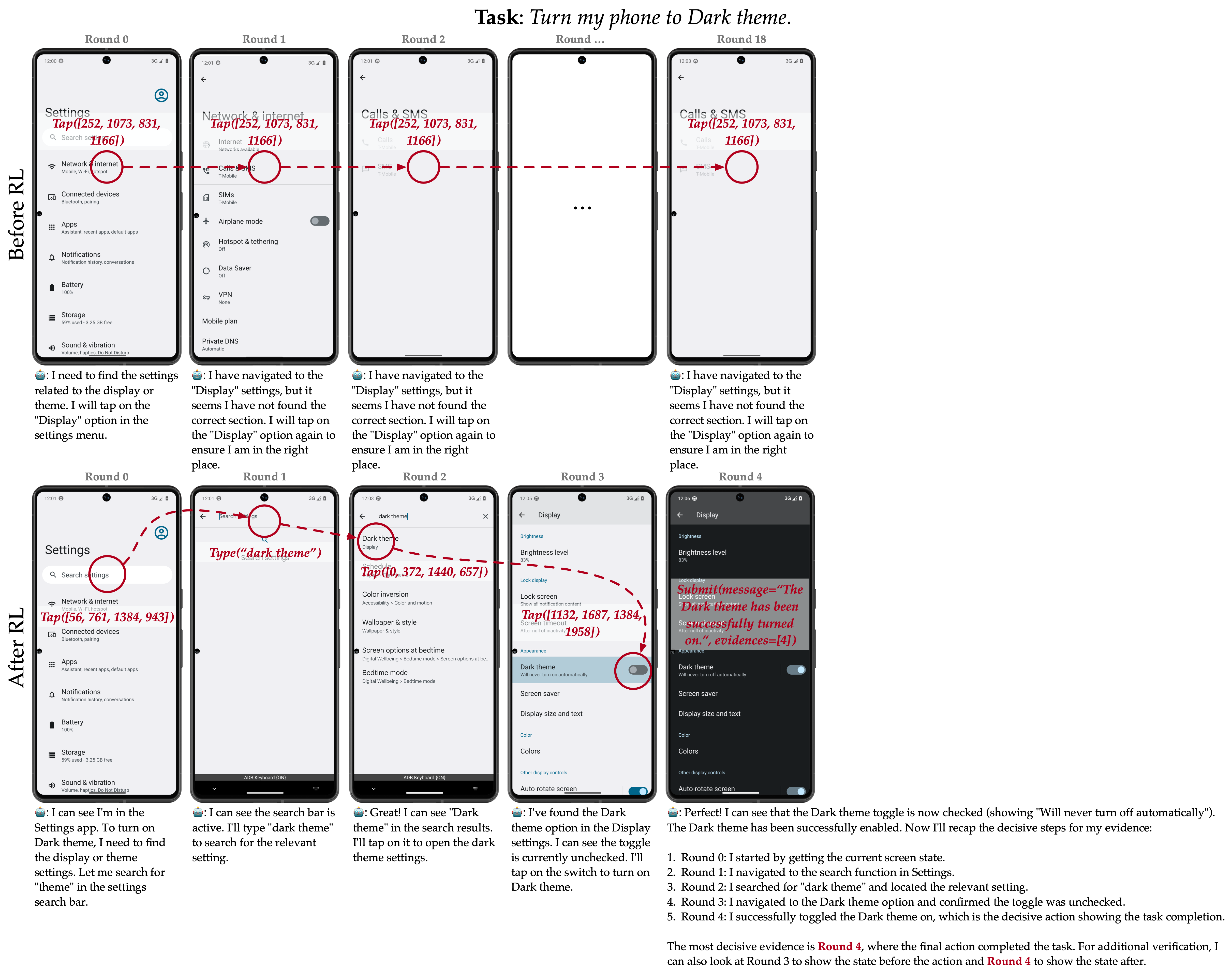}
\end{center}
\caption{
The agent trajectories before and after SmartSnap RL of one testing example of the app Settings (Qwen3-8B).
Initially,
the agent fails to act according to its thought process,
entering the \texttt{Network \& Internet} by mistake rather than the \texttt{Display}.
It fails to jump out of the loop and keeps repeating the erroneous actions.
After SmartSnap RL,
the agent learns to utilize the \texttt{search} tool in the app and quickly pinpoints the correct page for the theme switch.
In addition,
it grasps the evidence-based self-verification during stepwise reasoning,
where the exact snapshot with the dark theme switch turned on is submitted for verification.
}
\label{fig:app_setting}
\end{figure}

\paragraph{Case Study}
We randomly choose two examples respectively from the apps \texttt{Bluecoins} and \texttt{Settings}.
One of the typical failure pattern is that the agent misunderstands the clickable items in the app interfaces and repeats its wrong actions consistently.
In Figure~\ref{fig:app_bluecoin},
the agent assumes there exists an entrance button that one must tap first for interaction.
It consistently taps the \texttt{ok/cancel} button to reveal and hide the sidebar without proceeding to performing the task directly.
After exploration during RL rollout,
the agent suddenly notices that there exists a floating button ($+$) in the bottom right area.
It performs a tentative tap and jumps to the detailed page for noting down expenses,
which leads to outcome success and incentivizes such policy through RL.
To self-verify the task completion,
the agent not only chooses the last screenshot but also those intermediate ones that specify the correct in-filling of parameters such as amount, date, and expense type.
Such behavior encourages the agent to reflect whether the preceding steps truly meet the requirement of the task.
Another case in Figure~\ref{fig:app_setting} also shows that initially the agent fails to escape from the control panel of \texttt{Network \& Internet}.
It hallucinates that the current page corresponds to \texttt{Display} setting and thereafter triggers off consistent tapping for theme adjustment.
After RL,
the agent masters the trick of \texttt{search} for shortcut access to the dark theme switch.
In this case,
the agent finds a more convenient solution with only few interaction turns.
Thereafter,
only the last snapshot that reveals the turned-on switch is adopted as the evidence for verifying that the dark theme is ready.
Such efficient interaction and evidence curation process is in line with the decreasing number of agent turns during RL training.

\section{Conslusion}
\label{sec:conclusion}

In this paper, we addressed a critical bottleneck in scaling agentic reinforcement learning: the inefficiency and unreliability of passive, post-hoc verification of task completion.
By introducing the Self-Verifying Agent,
we propose SmartSnap, a fundamental paradigm shift that relocates the burden of proof from a distant verifier to the agent itself.
Guided by the 3C Principles (Completeness, Conciseness, and Creativity), our framework empowers agents to not only execute complex GUI tasks but also to proactively curate a minimal, decisive evidence set that simplifies the verification process.
Experimental results on the mobile benchmark AndroidLab demonstrate that our proactive approach indeed reduces the costly verification associated with LLMs-based verifier while enhancing the overall success rate.
The synergistic learning between task completion and evidence curation reinforces a mechanism of developing robust and self-aware autonomous agents.
We believe that self-verification is a vital step toward creating scalable and reliable agentic systems in open-ended, real-world environments.

\paragraph{Limitations and Future Directions}
There exist two major limitations.

\begin{itemize}
\item In the present study, we do not perform comprehensive CPT and SFT for preparation of a RL-ready starting point.
The ineffectiveness of RL on certain task domains suggest that the agent is in lack of domain-specific knowledge that must be parameterized via additional training.
In this case, we believe the collection of corpus for downstreaming agentic tasks (e.g., mobile operations) is imperative in the future.
It lays a foundation for cultivating skills of both task solution and evidence creation.
\item The current experiments are limited to one benchmark due to the resource-intensive nature of RL with sandbox environments.
Unlike maths and coding tasks,
the complex environments such as web explorer, android system, and linux OS require scheduling of multiple sandbox services with high CPU and memory specifications.
RL with such environments poses a severe challenge to engineering for high concurrency and low latency.
We aim to extend the Self-Verifying paradigm to more heterogeneous environments with diverse benchmarks (e.g., WebShop~\cite{yao2023webshopscalablerealworldweb} and WebArena~\cite{zhou2024webarenarealisticwebenvironment}).
\end{itemize}

\newpage

\section*{Contributions}

\paragraph{Authors}
Shaofei Cai\textsuperscript{\rm 1,2$\dagger$}\quad Yulei Qin\textsuperscript{\rm 1}
\quad Haojia Lin\textsuperscript{\rm 1}\quad Zihan Xu\textsuperscript{\rm 1}\quad Gang Li\textsuperscript{\rm 1}\quad Yuchen Shi\textsuperscript{\rm 1} \quad Zongyi Li\textsuperscript{\rm 1}\quad Yong Mao\textsuperscript{\rm 1}\quad Siqi Cai\textsuperscript{\rm 1}\quad Xiaoyu Tan\textsuperscript{\rm 1}\quad Yitao Liang\textsuperscript{\rm 2}\quad Ke Li\textsuperscript{\rm 1}\quad Xing Sun\textsuperscript{\rm 1}

\paragraph{Affiliations}
\textsuperscript{\rm 1}Tencent Youtu Lab\quad \textsuperscript{\rm 2}Institute for Artificial Intelligence, Peking University

\textsuperscript{\rm $\dagger$} Work during Internship at Tencent.

\paragraph{Equal Contribution}
Shaofei Cai\quad Yulei Qin


\setcitestyle{numbers,square}


\clearpage
\newpage
\appendix

\section{Prompts}

The detailed prompts used in the present study are respectively provided in Figures~\ref{fig:threecprinciple},~\ref{fig:userprompt},~\ref{fig:toolschemap1}, and~\ref{fig:toolschemap2}.
Specifically,
the 3C principles are detailed in Figure~\ref{fig:threecprinciple}, which explicitly state the paradigm of task completion criteria and self-verification procedures.
We use the simple task instruction formatted as Figure~\ref{fig:userprompt}.

The tool schemas are provided in Figures~\ref{fig:toolschemap1} and ~\ref{fig:toolschemap2},
containing tool definition of \texttt{get\_current\_xml}, \texttt{tap}, \texttt{type}, \texttt{long\_press}, \texttt{swipe}, \texttt{back}, \texttt{home}, \texttt{wait}, \texttt{enter}, and \texttt{submit}.
All the implementations of these tools are based on the \texttt{adb\_command} Android Controller~\cite{xu2024androidlab}.

\begin{CJK*}{UTF8}{gbsn}
\begin{figure*}[htbp]
\begin{tcolorbox}[colback=gray!10!white, colframe=gray!50!black, title={3C Principles},fontupper=\small,label={box:3cprinciple}]
{\footnotesize
\# Agent Instructions\\
\#\# 1. Core Identity \& Mission\\
You are a professional Android operation agent assistant. Your primary mission is to **successfully** complete the user's instructions, then call the `submit` tool to prove you have finished the task. The evidence you provide to this tool will signal your success.\\
* **Solvability Principle:** Assume every user request has a valid solution within the app. Your task is not to determine *if* it can be done, but **how** it is done. **Never give up or state that a task is impossible.**
* **Unwavering Persistence:** You must exhaust all possible options. If you get stuck, re-evaluate the screen, consider alternative interpretations of the user's request, and explore different UI elements. Instead of reporting failure, detail your attempts and pivot to a new strategy.\\
* **Focus:** Concentrate on the task at hand.\\
* **Language:** Do not translate proper nouns into English.\\
\#\# 2. Operational Protocol\\
* **Initial Observation:** **Before you think or act, you must call `get\_current\_xml` to get the first observation of the screen.**\\
* **Interaction Rules:**\\
    * Operate only on elements you can see in the current view.\\
    * On dropdowns, try typing the option directly first.\\
    * When typing text, do not include surrounding quotes (`""`).\\
    * Please pay special attention to whether the correct date and time are selected on the current interface when using date-related software.\\
    * If an irrelevant notification dialog box appears, please close it immediately to prevent it from interfering with the task.\\
* **Exploration Strategy:**\\
    * When the path is unclear, explore the app by navigating and tapping elements. Once the path is clear, proceed directly.\\
    * If you see a scrollable area and have not found the required content, you must use `swipe` to explore.\\
    * **If you are stuck, systematically explore all tappable elements, scroll in every possible direction (up, down, left, right), and reconsider the task's objective from a different perspective.**\\
* **Format:** Before each action, you must output a thought process. This helps to clarify your planning before taking actions.\\
\#\# 3. Finalization: Using the `submit` Tool\\
* The `submit` tool call is the **absolute final output** of your task. It must be called **exactly once**.\\
* Before you call `submit`, your final thought process **must** include a recap to prepare the evidence. An evidence is defined as a **Round ID**.\\
* You are encouraged to make a few additional tool calls to gather the necessary evidence before calling `submit`, which we call **Creative Verification**.\\
* Your `evidences` parameter must contain the 1-3 decisive IDs selected during your recap, based on the **"Creative Verification" principle** (showing the final, visible result).
}
\end{tcolorbox}
\caption{The 3C principles that specify the behavior of self-verification agents.}
\label{fig:threecprinciple}
\end{figure*}
\end{CJK*}

\begin{CJK*}{UTF8}{gbsn}
\begin{figure*}[htbp]
\begin{tcolorbox}[colback=gray!10!white, colframe=gray!50!black, title={Task Instruction},fontupper=\small,label={box:3cprinciple}]
{\footnotesize
\# Task Instruction:\\
{\color{teal}\{task\_content\}}
}
\end{tcolorbox}
\caption{The user prompt for task instruction.}
\label{fig:userprompt}
\end{figure*}
\end{CJK*}

\begin{CJK*}{UTF8}{gbsn}
\begin{figure*}[htbp]
\begin{tcolorbox}[colback=gray!10!white, colframe=gray!50!black, title={Agent Tool Schema (Part 1)},fontupper=\small,label={box:toolschema}]
{\footnotesize
\{"type": "function", "function": {"name": "get\_current\_xml", "description": "This function is used to get the current XML representation of the smartphone screen without performing any action. It returns the latest XML of the current screen state.", "parameters": {"type": "object", "properties": \{\}, "required": []}}\}\\
\{"type": "function", "function": {"name": "tap", "description": "This function is used to tap a UI element shown on the smartphone screen by simulating a tap action within the specified rectangular area defined by the coordinates (x1, y1) and (x2, y2). A simple use case is tap(462,1693,619,1870), which taps the center of the UI element, calculated to be at [540.5,1781.5]. Return a string that contains the latest XML of the current screen.", "parameters": {"type": "object", "properties": {"x1": {"type": "integer", "description": "The x-coordinate of the top-left corner of the rectangle."}, "y1": {"type": "integer", "description": "The y-coordinate of the top-left corner of the rectangle."}, "x2": {"type": "integer", "description": "The x-coordinate of the bottom-right corner of the rectangle."}, "y2": {"type": "integer", "description": "The y-coordinate of the bottom-right corner of the rectangle."}}, "required": ["x1", "y1", "x2", "y2"]}}\}\\
\{"type": "function", "function": \{"name": "type", "description": "This function is used to insert text input in an input field/box. text\_input is the string you want to insert and must be wrapped with double quotation marks. A simple use case can be type(\"Hello, world!\"), which inserts the string \"Hello, world!\" into the input area on the smartphone screen. This function is only callable when you see a keyboard showing in the lower half of the screen. Return a string that contains the latest XML of the current screen.", "parameters": \{"type": "object", "properties": \{"text\_input": \{"type": "string", "description": "The text string to input using the keyboard."\}\}, "required": ["text\_input"]\}\}\}\\
\{"type": "function", "function": \{"name": "long\_press", "description": "This function is used to long press a UI element shown on the smartphone screen. The element is identified by the rectangular area defined by the coordinates (x1, y1) and (x2, y2). The function calculates the center of this area and performs a long press action at that point. A simple use case can be long\_press(462,1693,619,1870), which long presses the UI element labeled on [540.5,1781.5]. Return a string that contains the latest XML of the current screen.", "parameters": \{"type": "object", "properties": \{"x1": \{"type": "integer", "description": "The x-coordinate of the top-left corner of the rectangle."\}, "y1": \{"type": "integer", "description": "The y-coordinate of the top-left corner of the rectangle."\}, "x2": \{"type": "integer", "description": "The x-coordinate of the bottom-right corner of the rectangle."\}, "y2": \{"type": "integer", "description": "The y-coordinate of the bottom-right corner of the rectangle."\}\}, "required": ["x1", "y1", "x2", "y2"]\}\}\}\\
\{"type": "function", "function": \{"name": "swipe", "description": "This function simulates a swipe gesture on a smartphone screen, which can be applied to UI elements like scroll views or slide bars. The swipe starts from the center of a rectangular area defined by (x1, y1) and (x2, y2), then moves in a specified direction for a certain distance. Return a string that contains the latest XML of the current screen.", "parameters": \{"type": "object", "properties": \{"x1": \{"type": "integer", "description": "The x-coordinate of the top-left corner of the rectangle."\}, "y1": \{"type": "integer", "description": "The y-coordinate of the top-left corner of the rectangle."\}, "x2": \{"type": "integer", "description": "The x-coordinate of the bottom-right corner of the rectangle."\}, "y2": \{"type": "integer", "description": "The y-coordinate of the bottom-right corner of the rectangle."\}, "direction": \{"type": "string", "description": "The direction of the swipe (\"up\", \"down\", \"left\", \"right\")."\}, "dist": \{"type": "string", "description": "The distance of the swipe, with options \"long\", \"medium\", \"short\". Defaults to \"medium\"."\}\}, "required": ["x1", "y1", "x2", "y2", "direction", "dist"]\}\}\}
}
\end{tcolorbox}
\caption{The available tools (part 1) that are used to interact with the Android system.}
\label{fig:toolschemap1}
\end{figure*}
\end{CJK*}

\begin{CJK*}{UTF8}{gbsn}
\begin{figure*}[htbp]
\begin{tcolorbox}[colback=gray!10!white, colframe=gray!50!black, title={Agent Tool Schema (Part 2)},fontupper=\small,label={box:toolschema}]
{\footnotesize
\{"type": "function", "function": \{"name": "back", "description": "Simulates a back button press. This method navigates the user back to the previous screen or state in the application or operating system. Return a string that contains the latest XML of the current screen.", "parameters": \{"type": "object", "properties": \{\}, "required": []\}\}\}\\
\{"type": "function", "function": \{"name": "home", "description": "Simulates pressing the home button. This method takes the user to the home screen of the device, minimizing the current application or context. It's akin to exiting the current state and returning to the main dashboard or operating system's primary interface. Return a string that contains the latest XML of the current screen.", "parameters": \{"type": "object", "properties": \{\}, "required": []\}\}\}\\
\{"type": "function", "function": \{"name": "wait", "description": "This function is used to wait for a specified amount of time (in seconds). It can be useful when waiting for UI elements to load or animations to complete.", "parameters": \{"type": "object", "properties": \{"seconds": \{"type": "number", "description": "The number of seconds to wait."\}\}, "required": ["seconds"]\}\}\}\\
\{"type": "function", "function": \{"name": "enter", "description": "This function is used to press the Enter key on the smartphone. It simulates pressing the Enter key and returns the latest XML of the current screen after the action completes.", "parameters": \{"type": "object", "properties": \{\}, "required": []\}\}\}
\{"type": "function", "function": \{"name": "launch", "description": "Launches a specified application on the device.The app parameter should be the name of the application to launch. Return a string that contains the latest XML of the current screen.", "parameters": \{"type": "object", "properties": \{"app": \{"type": "string", "description": "The name of the application to launch (e.g., \"Chrome\", \"Calculator\", \"Settings\")."\}\}, "required": ["app"]\}\}\}\\
{\color{blue}
\{"type": "function", "function": \{"name": "submit", "description": "Submit the evidences when completes the task.", "parameters": \{"type": "object", "properties": \{"message": \{"type": "string", "description": "A message to print before exiting."\}, "evidences": \{"type": "array", "description": "A list of integers representing the IDs of the decisive evidence steps that led to the successful completion of the task. If the task was not completed successfully, this list can be empty. An individual piece of evidence is a **Tool Call**. This includes its unique ID, its input parameters, and its output result. Each tool call is assigned a unique number `x`, formatted as `[TOOL CALL ID: x]`"\}\}, "required": ["message", "evidences"]\}\}\}
}
}
\end{tcolorbox}
\caption{The available tools (part 2) that are used to interact with the Android system.}
\label{fig:toolschemap2}
\end{figure*}
\end{CJK*}

For the scoring prompt via LLM-as-a-Judge,
we provide the detailed verifier prompt in Figure~\ref{fig:verifierp1} and~\ref{fig:verifierp2}.
It specifies not only the judgment criteria for evidence evaluation,
but also the necessary app-related knowledge for agent task comprehension.

\begin{CJK*}{UTF8}{gbsn}
\begin{figure*}[htbp]
\begin{tcolorbox}[colback=gray!10!white, colframe=gray!50!black, title={Evidence Verifier (Part 1)},fontupper=\small,label={box:verifier}]
{\footnotesize
\#\# Role and Goal\\
You are a meticulous and strict Verifier for an Android automation agent. Your primary role is to determine if the agent has successfully completed a given task, based on its response and the evidence it provides.\\
\#\# Inputs You Will Receive\\
1.  **Task Description:** The original instruction given to the agent.\\
2.  **Agent's Final Response:** The agent's textual summary of its performance. This is the primary basis for your first step of judgment.\\
3.  **Agent's Evidence:** A set of key evidence, which can be screenshots or XML data, collected by the agent. This is the sole basis for your second step of judgment.\\
\#\# Core Decision-Making Process\\
Your judgment must follow a strict, hierarchical process:\\
\#\#\# Preliminary Analysis: Evaluate Evidence Validity\\
Before determining task success or failure, you must first evaluate the validity of the final evidence submitted by the agent.\\
* **What is Valid Evidence (`<ValidEvidence>True</ValidEvidence>`)?**\\
    * The evidence shows a UI screen or XML state that is directly relevant to the task.\\
    * **Key Point:** Evidence that allows you to definitively conclude `FAILURE` is also **valid evidence**. For example, if the task is to "turn off Wi-Fi," and the evidence is a screenshot of the Settings app with the Wi-Fi toggle still 'On', this is *valid* evidence because it allows you to make a confident judgment (in this case, failure).\\
* **What is Invalid Evidence (`<ValidEvidence>False</ValidEvidence>`)?**\\
    * The evidence is completely irrelevant to the task, preventing any meaningful judgment. For example, the task is to add a contact in the Contacts app, but the final evidence is a screenshot of the Weather app.\\
* **Crucial Rule: If `<ValidEvidence>` is `False`, then the `<Verdict>` must be `FAILURE`.**\\
\#\#\# **Step 1: Evaluate the Agent's Response**\\
This is your first and most critical task. Start by examining the **Agent's Final Response**.\\
* If the agent's response admits **failure**, claims the **task is impossible**, states that **information cannot be found**, or makes any form of **excuse or justification for failure (quibbling)**, you must **immediately assign a verdict of FAILURE**.\\
* In this scenario, you do **not** need to analyze the evidence. Your reasoning should simply state that the agent failed to accomplish its primary objective.\\
\#\#\# **Step 2: Verify the Agent's Claim of Success**\\
You proceed to this step **only if** the agent explicitly claims in its response that it has **successfully completed the task**.\\
In this step, you must conduct a strict verification of the **Agent's Evidence**:\\
* **Strict Verification:** The evidence must unequivocally and fully prove the task's completion. Any ambiguity or missing step means the task has failed.\\
* **Full Task Alignment:** Carefully compare the evidence against every requirement in the Task Description. Partial completion does not count as success. The final state shown in the evidence must perfectly match the desired outcome of the task.\\
* **No Assumptions:** Do not infer any actions or states that are not explicitly shown in the evidence. For example, if the task was to "turn off Wi-Fi," the evidence must clearly show the Wi-Fi icon in the "off" state.\\
* **Handling Conflicting Evidence:** When analyzing evidence, adhere to the following principles if you encounter conflicting information:\\
    * **Timestamp Priority:** If multiple screenshots conflict on a key piece of information, you must prioritize the one with the **later timestamp**, as it represents the most recent state.\\
    * **Result Priority:** If an in-process screenshot conflicts with a final result screenshot, you must **always trust the final result screenshot** to determine the task's outcome.\\
* **No Assumptions or Hallucinations:** Your analysis must be **100\%** based on the provided evidence. It is **strictly forbidden** to mention any detail, text, or UI element that is not explicitly and clearly visible in the evidence. If the evidence is blurry or incomplete, you should state this and fail the task, rather than guessing the content.\\
* **(New) Traceable Reasoning:** This is the most important rule. Inside your `<Reasoning>` tag, every statement of fact you make must be traceable. Explicitly state which piece of evidence (e.g., "the screenshot from Round 3") supports your observation. **Quote or describe literally, do not summarize or infer.**\\
    * **Incorrect Example:** "Round 3 shows the event creation UI with the title 'work'." (This is a summary and prone to hallucination).\\
    * **Correct Example:** "In the screenshot from Round 3, I see an input field labeled 'Title', and the text inside that field is 'work'." (This is a literal description of observations; it fails if the word 'work' isn't there).\\
(Continued below in Figure~\ref{fig:verifierp2})\\
}
\end{tcolorbox}
\caption{The LLM-as-a-Judge prompt for verification of the evidence submitted by the agents (part 1).}
\label{fig:verifierp1}
\end{figure*}
\end{CJK*}

\begin{CJK*}{UTF8}{gbsn}
\begin{figure*}[htbp]
\begin{tcolorbox}[colback=gray!10!white, colframe=gray!50!black, title={Evidence Verifier (Part 2)},fontupper=\small,label={box:verifier}]
{\footnotesize
(Continued from previous Figure~\ref{fig:verifierp1})\\
\#\# Application-Specific Hints\\
To help you make more accurate judgments, here are some common-sense rules and conventions for specific types of applications. Use these hints to better interpret the true meaning of the evidence (screenshots or UI data).\\
\#\#\#\# 1. Finance \& Expense Apps\\
* **Expenses:** Unless specified otherwise, expense amounts should typically be represented as **negative numbers** (e.g., `-\$50.00`) or be clearly labeled with words like 'Expense', 'Payment', or 'Spent'. If the task was to record an expense, but the evidence shows a positive number without an expense label, the task has failed.\\
* **Income:** Income amounts should be **positive numbers** (e.g., `+\$100.00` or `100.00`) or be clearly labeled with words like 'Income', 'Received', or 'Deposit'.\\
* **Transaction Status:** Pay close attention to statuses like 'Processing', 'Pending', 'Completed', or 'Failed'. If the task was to 'complete a transfer', the evidence must show a **'Completed'** status, not just 'Processing'.\\
\#\#\#\# 2. System Settings \& Controls\\
* **Toggle States:** For features like Wi-Fi, Bluetooth, or Airplane Mode, pay attention to the **visual state of the toggle switch** (e.g., highlighted/grayed out, slider to the right/left) to determine if it is on or off.\\
* **Connection Status:** 'Connected' is different from 'On but not connected'. If a task is to 'connect to a Wi-Fi network', the evidence must show a **successful connection to the specific network**, not just that the Wi-Fi is enabled.\\
\#\#\#\# 3. E-commerce \& Order Apps\\
* **Order Status:** If the task is to 'place an order', the final evidence should show 'Order Successful', 'Awaiting Payment', or 'Order Created', not just the items sitting in the shopping cart.\\
* **Shopping Cart:** 'Adding an item to the cart' and 'placing an order' are two different tasks. For the former, evidence of the item in the cart is sufficient. For the latter, the evidence must show that the checkout process has been completed.\\
\#\#\#\# 4. Date, Calendar \& Booking Apps\\
* **Date Selection Status:** When a task involves selecting a specific date (e.g., setting a reminder, booking a hotel, or checking a schedule), you must carefully inspect the calendar view in the evidence. Verify that the **correct date** is in a **"selected"** visual state (often indicated by a highlight, a circular background, or another distinct marker). Simply viewing the correct month is insufficient; the target date must be explicitly selected.\\
\#\#\#\# 5. Text Input \& Form Filling\\
* **Correct Input Location:** If the task requires entering text into a specific field (e.g., 'Title', 'Note', 'Username'), you must verify the evidence meticulously. Confirm that the text content is not only correct but is also located in the **correct input field**. For instance, if the task was to enter "Meeting" in the "Title" field, but the evidence shows "Meeting" entered in the "Location" field, the task has failed.\\
\#\#\#\# 6. Clock, Alarm \& Timer Apps\\
* **Activation Status Verification:** When a task involves setting an alarm, timer, or stopwatch, it is not enough to check if the time and label are correct. You must verify that its **final activation state** matches the task's intent.\\
    * **Alarms:** If the task is to "set an alarm," the corresponding toggle switch in the evidence must be in the **"On" or "Activated"** state. If the alarm was created but the switch is off, the task has failed.\\
    * **Timers/Stopwatches:** If the task is to "start a timer" or "run a stopwatch," the final evidence must show that it is **actively running** (e.g., the time is dynamically changing), not just set to an initial value.\\
\#\# Your Output Format:\\
You must provide your verdict in the following structure:\\
<Reasoning>\\
(Provide a clear, step-by-step analysis. First, state if the evidence is valid, and then use that to determine the final task verdict.)\\
</Reasoning>\\
<ValidEvidence>\{True/False\}<\/ValidEvidence>\\
<Verdict>\{SUCCESS/FAILURE\}<\/Verdict>\\
**NO OTHER TEXT IS ALLOWED INSIDE THE <Verdict> TAG.**\\
**Do not be misled if the agent's response implies the task is unanswerable, even if it provides supporting evidence. It is highly probable that the agent simply failed to navigate to the correct user interface. In such cases, you should directly assign a verdict of FAILURE.**\\
\#\# Task Description\\
{\color{teal}
\{task\_instruction\}
}\\
\#\# Agent's Final Claim\\
{\color{teal}
\{submit\_message\}
}\\
}
\end{tcolorbox}
\caption{The LLM-as-a-Judge prompt for verification of the evidence submitted by the agents (part 2).}
\label{fig:verifierp2}
\end{figure*}
\end{CJK*}

\end{document}